\documentclass[11pt]{article}

\usepackage[final]{acl}

\usepackage{times}
\usepackage{latexsym}

\usepackage[T1]{fontenc}

\usepackage[utf8]{inputenc}

\usepackage{microtype}

\usepackage{inconsolata}

\usepackage{graphicx}
\usepackage{svg}
\usepackage{amsmath}
\usepackage{enumitem}
\usepackage{caption}
\usepackage{multirow} 
\usepackage{wrapfig}
\usepackage[skins,breakable,theorems]{tcolorbox}
\usepackage{inconsolata}
\usepackage{listings}
\usepackage{xcolor}
\usepackage{subcaption}
\usepackage{placeins}
\newcommand{\promptsubsection}[1]{
\setlength{\parskip}{6pt} \noindent\textbf{{#1}:}
}
\usepackage{makecell}
\usepackage{tabularx}
\usepackage{lineno}
\usepackage{booktabs}
\usepackage{fvextra}

\DefineVerbatimEnvironment{PromptVerbatim}{Verbatim}{
  breaklines=true,
  breakanywhere=true,
  fontsize=\scriptsize,
  baselinestretch=0.95
}

\definecolor{darkblue}{rgb}{0, 0, 0.5}
\hypersetup{colorlinks=true, citecolor=darkblue, linkcolor=darkblue, urlcolor=darkblue}

\title{Evaluating Harmful Memorization in LLM Code Generation}

\author{
Lizhe Zhang\thanks{Equal contribution.},
Wentao Chen\footnotemark[1],
Li Zhong,
Letian Peng,
Zilong Wang,
Jingbo Shang \\
University of California, San Diego \\
\texttt{\{liz058,wec056,lizhong,lepeng,zlwang,jshang\}@ucsd.edu}
}

\begin{document}
\maketitle
\begin{abstract}

Assessing memorization in LLM code generation is challenging because code reuse is not inherently harmful. Correct programs often share APIs, templates, and algorithmic idioms, so code similarity alone may reflect benign reuse rather than problematic memorization. We focus on the harmful case: a model fails to adapt when task semantics change and instead produces code that remains highly similar to the original solution. We propose \emph{code rewriting}, an automated semantic perturbation pipeline that constructs superficially similar but semantically different variants of existing coding tasks by modifying a core logic element of the reference solution and regenerating the corresponding prompt. To summarize this behavior, we introduce the \emph{Harmful Memorization Index} (HMI), an operational diagnostic that combines similarity to the original solution with performance degradation on rewritten tasks. We use harmful memorization as an operational behavioral term for this specific failure mode; HMI does not establish whether an original task or solution appeared in a model's training data. HMI is intended to complement standard accuracy metrics by capturing brittleness under small semantic changes. Across MBPP+ and BigCodeBench, we find that: (1) HMI does not monotonically increase with model scale and often decreases on simpler tasks; (2) high-HMI failures are concentrated in specific logic-edit types, especially constraint changes and data transformations; (3) supervised fine-tuning improves original-task accuracy but can substantially increase HMI on held-out tasks; and (4) across the evaluated post-training settings, PPO yields a more favorable accuracy--HMI trade-off than SFT.

\end{abstract}

\section{Introduction}

Large language models (LLMs) have made rapid progress in code generation and are increasingly used as practical tools in software development~\citep{sourcegraphcody, tabnine2023, team2023gemini, anthropic2025claudecode, humaneval}. Modern code-focused LLMs achieve strong performance on programming benchmarks~\citep{rozière2024codellamaopenfoundation, hui2024qwen25codertechnicalreport}, making benchmark accuracy an essential measure of code-generation capability. However, accuracy alone does not reveal how a model behaves when a familiar-looking programming task requires a different solution.

This distinction is important because memorization in code generation is not uniformly harmful. Many correct programs reuse common APIs, boilerplate, library conventions, programming idioms, or canonical algorithmic templates. Penalizing code similarity alone would therefore mischaracterize useful knowledge reuse as undesirable memorization. The harmful case is more specific: a model may continue to rely on a familiar solution pattern even after the task semantics have changed, producing code that resembles the original solution but no longer satisfies the new requirement. In this setting, the key question is not whether the generated code is similar to existing code, but whether such similarity remains semantically appropriate.

\definecolor{par_mut}{HTML}{6C8EBF}
\definecolor{rew}{HTML}{D6B656}
\definecolor{prompt}{HTML}{AA6033}
\begin{figure*}[t]
    \centering
    \includegraphics[width=\textwidth]{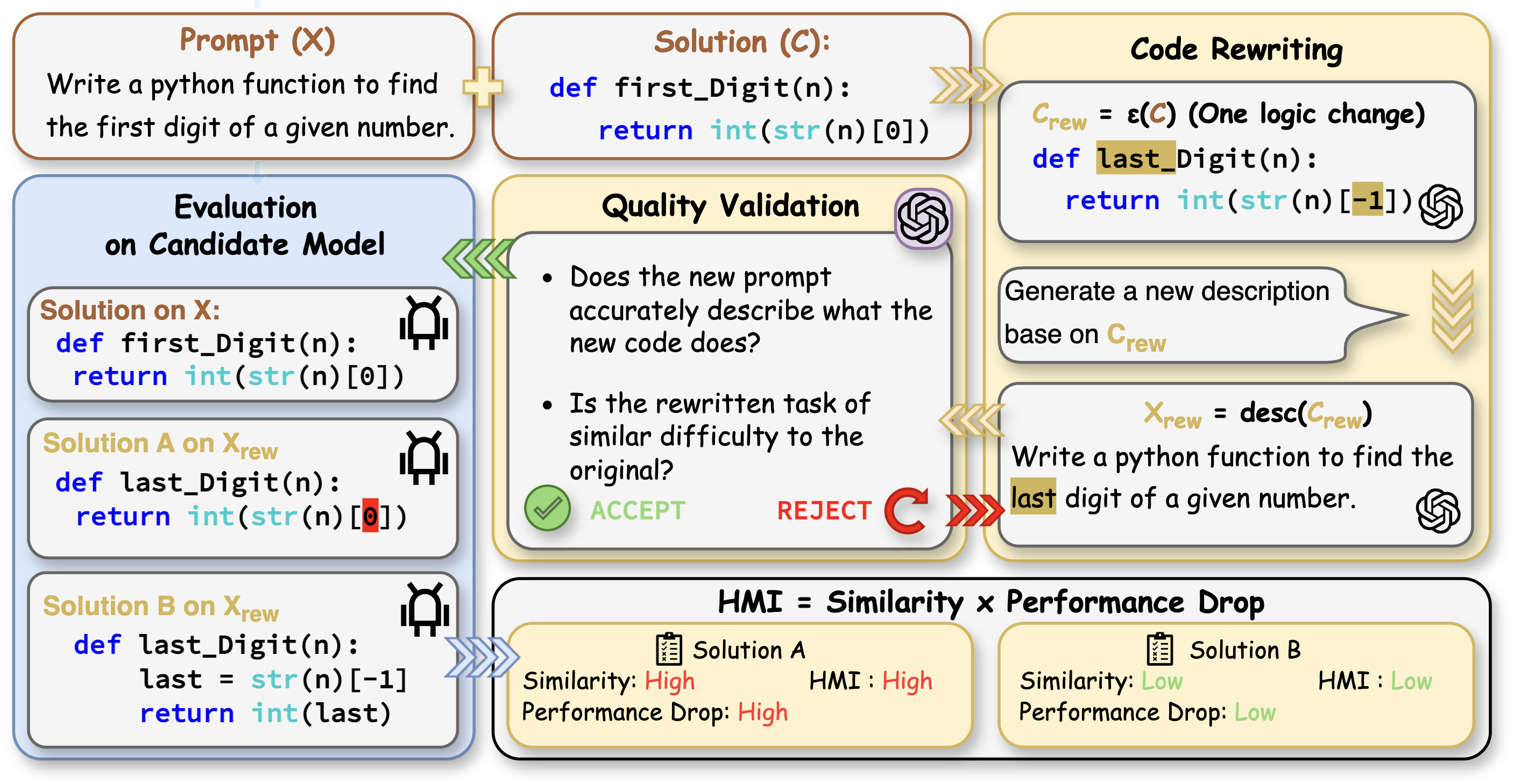}
    \caption{Overview of the automated \textcolor{rew}{code rewriting} and HMI evaluation pipeline. \textcolor{rew}{Code rewriting} first changes a single logic element within the ground-truth solution \textcolor{prompt}{C} while preserving the original function's structure. It then generates a new prompt \textcolor{rew}{$X_\text{rew}$} based on \textcolor{rew}{$C_\text{rew}$}. An LLM-as-a-judge verifies that \textcolor{rew}{$C_\text{rew}$} matches \textcolor{rew}{$X_\text{rew}$} and aligns with the original task's difficulty. Finally, the candidate model's output is evaluated using the Harmful Memorization Index (HMI), which combines the structural and semantic similarity of the model's answer with its relative performance drop on the rewritten task.}
    \label{fig:workflow_pipeline}
    \vspace{-0.8em}
\end{figure*}

We use the term \emph{harmful memorization} operationally to denote the following behavioral failure mode: a model succeeds on an original task but fails to adapt to a semantically rewritten variant while producing code that remains highly similar to the original solution. This behavioral definition does not claim that the original task or solution appeared in the model's training data. To operationalize this behavior, we introduce \emph{code rewriting}, an automated semantic perturbation pipeline that constructs near-neighbor coding tasks. Given an original task and its reference solution, the pipeline modifies one core logic element of the solution while preserving much of its surrounding structure, and then regenerates a corresponding task prompt. The resulting rewritten task is superficially similar to the original task but requires different underlying logic. This controlled setup allows us to ask: when the surface form remains familiar but the semantics change, does the model adapt its solution, or does it regress to the original solution pattern?

We summarize this behavior with the \emph{Harmful Memorization Index} (HMI), an operational diagnostic that combines two signals: (i) the similarity between the model's answer on the rewritten task and the original reference solution, and (ii) the relative performance drop from the original task to the rewritten task. HMI is most informative when interpreted alongside standard accuracy metrics. A model with high original-task accuracy and low HMI both solves the benchmark and adapts to semantic rewrites. In contrast, a model with high original-task accuracy and high HMI may perform well on the original benchmark while remaining brittle under small but consequential semantic changes. Low HMI should not by itself be interpreted as strong coding ability, since a model with low original-task accuracy may simply fail without reproducing the original solution pattern. More importantly, HMI is a diagnostic of an observable behavioral pattern rather than direct evidence of training-set memorization or its causal origin. High HMI may also arise from other forms of semantic adaptation failure, which we discuss in Section~\ref{sec:limitations}.

Our evaluation focuses on function-level coding benchmarks across different difficulty levels, from simpler problems in \textsc{MBPP+}~\citep{evalplus} to more difficult tasks in \textsc{BigCodeBench}~\citep{zhuo2024bigcodebench}. This scope enables controlled semantic rewrites and fine-grained attribution of failures to localized logic changes, such as constraint changes, data transformations, component swaps, and logic reversals. Using this setup, we evaluate multiple model families and post-training strategies. Fine-tuning is not required to compute HMI: we first apply the diagnostic directly to existing instruction-tuned and code-focused models. We then use Supervised Fine-Tuning (SFT) and Proximal Policy Optimization (PPO) as additional case studies to analyze how post-training changes the accuracy--HMI trade-off. Our study shows that HMI does not monotonically increase with scale, is concentrated in specific logic-edit types such as constraint changes and data transformations, and changes substantially under post-training, with SFT increasing HMI more than PPO.

In summary, our work makes the following contributions:
\begin{itemize}[leftmargin=*,nosep]
\item We formulate harmful memorization in code generation behaviorally as: failure on a semantically rewritten task under high similarity to the original solution, rather than similarity alone.
\item We propose an automated \textbf{code-rewriting diagnostic} that constructs superficially similar but semantically different variants of existing coding tasks by modifying localized solution logic.
\item We introduce HMI as an operational index for summarizing this failure mode jointly with original-task accuracy, similarity, and relative accuracy drop.
\item We conduct an empirical study across model scale, benchmark difficulty, logic-change types, and post-training strategies, showing when code models adapt to changed semantics and when they regress to familiar solution patterns.
\end{itemize}

\section{Related Work}

\paragraph{LLMs for Code Generation.}
Recent LLMs achieve strong results on code-generation benchmarks such as \textsc{BigCodeBench}, LiveCodeBench, and EvalPlus~\citep{zhuo2024bigcodebench,jain2024livecodebenchholisticcontaminationfree,evalplus}. These benchmarks provide increasingly rigorous tests of functional correctness, task diversity, and contamination-aware evaluation. Our work is complementary: rather than replacing functional correctness evaluation, we use semantic rewriting to analyze how models behave when tasks remain familiar but require different solutions.

\paragraph{Memorization and Code Reuse.}
Prior work studies memorization in language and code models through extraction attacks, prefix-to-suffix completion, clone detection, data contamination analysis, and overlap with training corpora~\citep{carlini2019secretsharerevaluatingtesting, pappu2024measuringmemorizationrlhfcode, duan2024uncovering, kassem2024alpaca, hartmann2023sokmemorizationgeneralpurposelarge, riddell2024quantifyingcontaminationevaluatingcode, dong2024generalizationmemorizationdatacontamination, xie2024memorizationinreasoning, lu2024scalinglawsfactmemorization, staab2023beyond, zanella2020analyzing, humaneval}. These studies show that models can reproduce or closely resemble seen content, and several works report that measured memorization can increase with model size~\citep{Yang_2024,Al_Kaswan_2024,wang2024unlockingmemorizationlargelanguage}. However, in code generation, similarity is difficult to interpret: correct programs often share APIs, boilerplate, library conventions, and canonical algorithmic structures. As a result, similarity-based memorization measures can conflate harmful rote reuse with benign code reuse that is necessary for correct programming.

Recent work has therefore emphasized that memorization should be interpreted through its effect on downstream behavior rather than through overlap alone~\citep{bayat202pitfallsmemorizationmemorizationhurts}. Closest to our setting, \citet{lai2022ds1000naturalreliablebenchmark} use semantic perturbations to probe whether models adapt when reference-solution semantics change without increasing task difficulty. Our work follows this behavioral view but focuses on general-purpose function-level code-generation benchmarks. We introduce an automated \emph{code-rewriting} diagnostic that constructs superficially similar but semantically different task variants, and we summarize model behavior with HMI, an operational index combining similarity to the original solution with performance degradation on the rewritten task. This allows us to analyze harmful memorization as failure under high similarity, while separating it from benign reuse where similar code remains correct.

\section{Methodology}

\subsection{Code Rewriting}
\label{sec:code rewriting}
To probe this behavioral failure mode, we use \emph{code rewriting} to generate problems that are \textbf{semantically different} from the original tasks while remaining superficially similar. 

This automated pipeline is shown in~\autoref{fig:workflow_pipeline}. Specifically, we first modify \textbf{one core logic element} in the reference solution while preserving the original function structure. We then generate a new task description that reflects the altered code while remaining syntactically similar to the original task. Formally, let $x \in X$ be the original prompt in text space $X$ and $c \in C$ its reference code solution in code space $C$. We apply a rewriting function $\epsilon$ that produces a rewritten solution $c_{\mathrm{rew}} = \epsilon(c)$, where $c_{\mathrm{rew}}$ is functionally different from $c$ but shares the same function signature. The new prompt $x_{\mathrm{rew}}$ is then generated from $c_{\mathrm{rew}}$:
\begin{align}
x_{\mathrm{rew}} &= \text{desc}(c_{\mathrm{rew}}), \\
\text{sig}(c_{\mathrm{rew}}) &= \text{sig}(c), \quad c_{\mathrm{rew}} \not\equiv c ,
\end{align}
where $\text{desc}(\cdot)$ generates a task description from code and $\text{sig}(\cdot)$ extracts the function signature.

\paragraph{Data Validation.}
We validate rewritten tasks with three checks. First, GPT-5~\citep{openai2025gpt5} verifies prompt--solution consistency and comparable difficulty; this LLM judge is used only for dataset validation, not for scoring candidate models. Second, each rewritten solution is executed against its corresponding rewritten unit tests (see~\autoref{Code Rewriting Solution Validation}). Third, four experienced Python programmers independently reviewed all rewritten pairs, blind to model outputs, accuracy, and HMI, for prompt--solution alignment and difficulty equivalence. Most rewrites were judged similar in difficulty (91\% on MBPP+ and 79\% on BigCodeBench), while the proportions judged harder and easier were nearly balanced (4\% vs. 5\% on MBPP+, and 11\% vs. 10\% on BigCodeBench). Full human-validation statistics and acceptance criteria are provided in~\autoref{appendix:validation}. Representative regressed examples are shown in~\autoref{appendix:regressed}.

\subsubsection{Harmful Memorization Index}
HMI consists of two signals: (i) how similar the model’s answer for the rewritten task is to the original ground-truth solution, and (ii) the extent of the performance drop from the original task to its rewritten counterpart.
\paragraph{(i) Similarity}
\label{par:similarity}
For every rewritten task \( i \in \mathcal{T}_{\mathrm{rew}} \), we measure how closely the model-generated solution for the rewritten prompt resembles the original reference solution. We use two complementary, deterministic similarity measures:
\begin{itemize}[leftmargin=*,nosep]
    \item \textbf{AST similarity}: normalized tree-edit overlap between abstract syntax trees, capturing structural resemblance.
    \item \textbf{Edit similarity}: $1 - \frac{\text{Levenshtein distance}}{\text{max length}}$ over tokenized code, capturing surface-form resemblance.
\end{itemize}
Formally, let \(\mathrm{AST}_i \in [0,1]\) denote AST similarity and let \(\mathrm{Edit}_i \in [0,1]\) denote edit similarity. We combine them into a unified similarity score:
\begin{align}
S_i = \frac{\mathrm{AST}_i + \mathrm{Edit}_i}{2}.
\end{align}

At the corpus level, we define the mean similarity over all rewritten tasks as:
\begin{align}
\text{Sim}(\mathcal{T}_{\text{rew}}) = \frac{1}{|\mathcal{T}_{\text{rew}}|}\sum_{i\in\mathcal{T}_{\text{rew}}} S_i, \quad \text{Sim}\in[0,1].
\end{align}

\paragraph{(ii) Relative Accuracy Drop}
\label{par:rad_rew}
For a task set $\mathcal{T}$, Pass@1 is reported as $\mathrm{Acc}(\mathcal{T})$. 
To capture the performance loss induced by semantic rewriting, we define
\begin{equation}
\mathrm{RAD}_{\text{rew}}
=
\max\!\left(
0,\;
1 -
\frac{\mathrm{Acc}(\mathcal{T}_{\mathrm{rew}})}
     {\mathrm{Acc}(\mathcal{T}_{\mathrm{ori}})}
\right),
\label{eq:rad_rew}
\end{equation}
$\mathrm{RAD}_{\text{rew}}=0$ when rewriting does not hurt accuracy and increases when it does; 
the $\max(0,\cdot)$ prevents negative values when the performance on rewritten tasks happens to be better.

\paragraph{Corpus-Level HMI and Instance-Level iHMI}
\label{HMI def}
We formulate the final index in two complementary ways. Our primary metric, the corpus-level $\text{HMI}$, is defined as the product of the mean similarity and the overall relative accuracy drop:
\begin{align}
\text{HMI} = \text{Sim}(\mathcal{T}_{\text{rew}})\;\times\;\mathrm{RAD}_{\text{rew}},
\quad \text{HMI}\in[0,1].
\end{align}
Unlike an additive design, this multiplicative design makes HMI high only when both diagnostic conditions hold: (i) the model output for the rewritten task remains highly similar to the original reference solution and (ii) performance drops on the rewritten tasks. HMI is therefore a compact summary of the conjunction of these two signals rather than an independent source of evidence. We interpret $\mathrm{Sim}(T_{\mathrm{rew}})$, $\mathrm{RAD}{\mathrm{rew}}$, and $\mathrm{Acc}(T{\mathrm{ori}})$ alongside HMI in our analyses. High similarity with low RAD is consistent with benign reuse of the original solution pattern, whereas high RAD with low similarity is more consistent with a generic failure under rewriting rather than the specific behavioral pattern targeted by HMI. High values of both similarity and RAD correspond to the high-similarity rewrite failure captured by HMI. Moreover, when original-task accuracy is low, HMI is inherently difficult to interpret because the model has not demonstrated sufficient competence on the original tasks.

While corpus-level HMI is convenient for comparing models, it combines corpus-level averages. We therefore also report an instance-level variant, iHMI, which directly couples similarity and failure on the same examples:
\begin{equation}
    \mathrm{iHMI} 
    = \frac{1}{|\mathcal{T}_{\mathrm{passed}}|} 
      \sum_{k \in \mathcal{T}_{\mathrm{passed}}} \bigl(S_k \cdot \mathrm{Drop}_k\bigr),
\end{equation}
where $\mathcal{T}_{\mathrm{passed}}$ is the set of tasks that the model solves correctly on the original benchmark, $S_k$ is the unified similarity score for example $k$, and $\mathrm{Drop}_k \in \{0,1\}$ indicates whether the model fails on the rewritten task. Conditioning on $\mathcal{T}_{\mathrm{passed}}$ reflects on regression under semantic rewriting: failure on a rewritten task is most informative when the model solved the corresponding original task.

\paragraph{Accounting for low baseline accuracy.}
Because $\mathrm{RAD}_{\mathrm{rew}}$ is normalized by original-task accuracy, it can be unstable when $\mathrm{Acc}(\mathcal{T}_{\mathrm{ori}})$ is very low. We therefore treat HMI as meaningful only for model--benchmark pairs with sufficient original-task competence. In our main comparisons, we report $\mathrm{RAD}_{\mathrm{rew}}$ and HMI when $\mathrm{Acc}(\mathcal{T}_{\mathrm{ori}}) \geq 10\%$ and the model solves at least 50 original tasks. For lower-competence settings, we still report $\mathrm{Acc}(\mathcal{T}_{\mathrm{ori}})$ and $\mathrm{Acc}(\mathcal{T}_{\mathrm{rew}})$, but mark $\mathrm{RAD}_{\mathrm{rew}}$ and HMI as ``N/A'' to avoid over-interpreting noisy relative drops. We provide additional justification in~\autoref{appendix:mri_stability}.

\begin{figure*}[t]
    \centering
    \includegraphics[width=\linewidth]{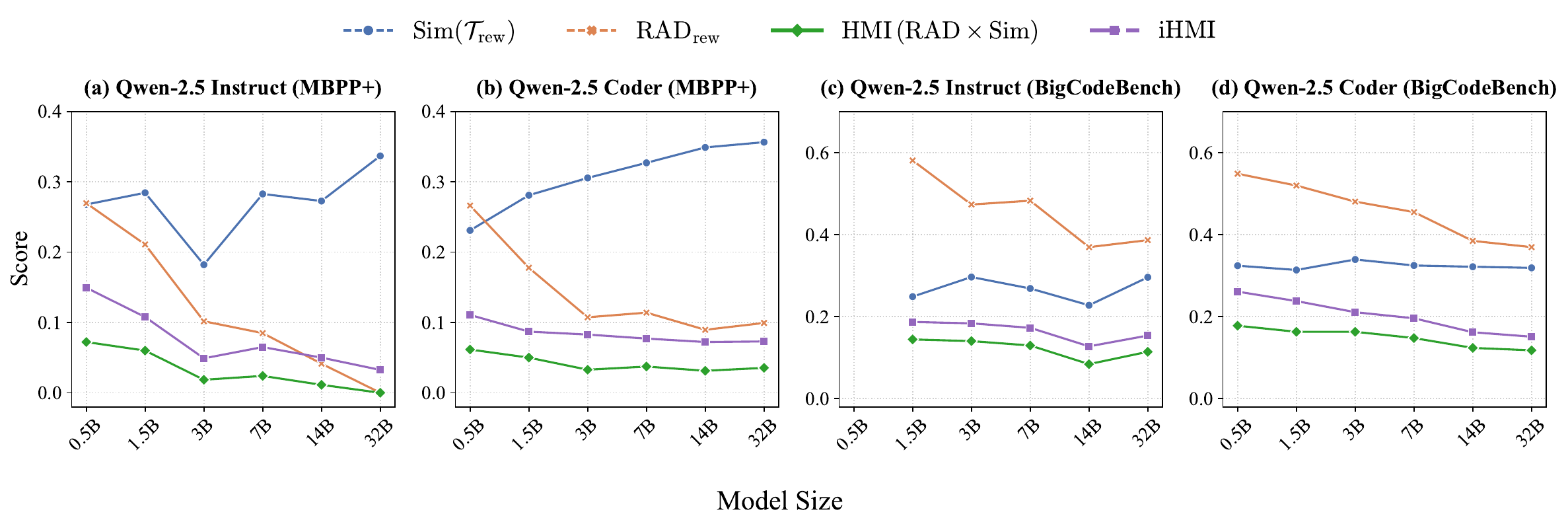}
    \caption{Scaling trends in HMI across Qwen-2.5 family. The Qwen-2.5 Instruct-0.5B on \textsc{BigCodeBench} is omitted because its baseline Pass@1 falls below our competence threshold (\autoref{appendix:mri_stability}).}
    \label{fig:mem-qwen-series}
\end{figure*}

\subsection{Post-Training Case Studies}
\label{finetuning_methods}
\paragraph{Supervised Fine-Tuning}
In SFT, coding problems are used as inputs and reference solutions as targets. We train models on the training split and evaluate HMI on the held-out test split. Since prolonged SFT can improve training-set fit while potentially reducing robustness to semantic rewrites, we evaluate multiple checkpoints around the validation-loss optimum. This allows us to examine whether additional accuracy gains are accompanied by higher HMI.

\paragraph{Reinforcement Learning}
We also consider PPO~\citep{schulman2017proximalpolicyoptimizationalgorithms} as an alternative post-training strategy. PPO optimizes a reward based on functional correctness of generated code while regularizing updates through a reference policy. We use the same model architecture for the actor, critic, and reference models, and select the checkpoint with the highest validation reward for evaluation. Comparing PPO with SFT allows us to analyze whether post-training methods that optimize executable correctness induce different accuracy--HMI trade-offs.

\section{Experiment Setup}

\subsection{Datasets}
\label{fine tune dataset details}
We conduct our evaluation on two widely-adopted code generation benchmarks: \textsc{MBPP+} \citep{evalplus} and \textsc{BigCodeBench}~\citep{zhuo2024bigcodebench}.
\paragraph{Dataset Statistics}
\textsc{MBPP+} contains 378 tasks, and \textsc{BigCodeBench} comprises 1140 tasks. We use a 4:1 train/test split for fine-tuning. For models without fine-tuning, we use the \textbf{complete set} of tasks for evaluation. For SFT and PPO, we train on the \textbf{training split} and evaluate on the \textbf{test split}\footnote{For clarity, both SFT and PPO are initialized from the same base model and trained independently; PPO is not performed on top of an SFT checkpoint.}. Due to the small size of \textsc{MBPP+} test split ($n=78$), estimates on this split may be noisy, we only use \textsc{BigCodeBench} to explore the impact of fine-tuning strategies on HMI.

\paragraph{Code Rewriting}
We generate one perturbed variant for each original task. More about the generation process is given in ~\autoref{appendix_dataset}.

\subsection{Models}
In this paper, we conduct our experiments on Qwen-2.5 series~\citep{hui2024qwen25codertechnicalreport}, Qwen-2.5-Coder series~\citep{qwen2025qwen25technicalreport}, Llama-3.1 series~\citep{dubey2024llama} and Llama-4 series~\citep{meta2025llama}. For fine-tuning, we choose Qwen-2.5-7B, and Qwen-2.5-Coder-7B. All training and inference were conducted on a server equipped with four NVIDIA A100 GPUs (80GB), with a total computational budget of approximately 40 GPU hours, using PyTorch and HuggingFace Transformers. More training details can be found at~\autoref{appendix:finetuning_details}.

\section{Results}
\subsection{Main results}


\begin{table}[h]
\centering
\small
\begin{tabular}{lrr}
\toprule
\textbf{Logic Change Type} & \textbf{BigCodeBench} & \textbf{MBPP+} \\
\midrule
Component Swap          & 393 (34.5\%) & 52 (13.8\%) \\
Data Transformation     & 385 (33.8\%) & 58 (15.3\%) \\
Constraint Change       & 185 (16.2\%) & 86 (22.8\%) \\
Logic Reversal          & 50 (4.4\%)   & 68 (18.0\%) \\
Workflow Modification   & 50 (4.4\%)   & 10 (2.6\%)  \\
Constant Swap           & 37 (3.3\%)   & 36 (9.5\%)  \\
Operation Order Swap    & 20 (1.8\%)   & 12 (3.2\%)  \\
Operator Change         & 18 (1.6\%)   & 36 (9.5\%)  \\
Other                   & 2 (0.2\%)    & 20 (5.3\%)  \\
\midrule
\textbf{Total}          & \textbf{1140 (100.0\%)} & \textbf{378 (100.0\%)} \\
\bottomrule
\end{tabular}
\caption{Statistics of Code Rewriting logic change types for BigCodeBench and MBPP+.}
\label{tab:logic-change-types-merged}
\vspace{-10pt}
\end{table}

\definecolor{High}{RGB}{220, 20, 60}   
\definecolor{Mid}{RGB}{255,140,0}     

\begin{table*}[t]
        \centering
        \renewcommand{\arraystretch}{1.2}
        \small
        \resizebox{\textwidth}{!}{
        \begin{tabular}{lrrrrr}
        \toprule
         & \multicolumn{5}{c}{\textbf{HMI on MBPP+}} \\
        \cmidrule(lr){2-6}
        \textbf{Model} 
        & Component Swap 
        & Constraint Change 
        & Data Transformation 
        & Logic Reversal 
        & \textbf{All} \\
        \midrule
        Qwen-2.5-0.5B-Instruct   & 0.0212 & 0.0929 & 0.0877 & \textcolor{High}{0.1119} & 0.0722 \\
        Qwen-2.5-1.5B-Instruct   & \textcolor{High}{0.0644} & 0.0584 & 0.0592 & 0.0553 & 0.0600 \\
        Qwen-2.5-3B-Instruct     & 0.0149 & \textcolor{High}{0.0380} & 0.0224 & 0.0000 & 0.0185 \\
        Qwen-2.5-7B-Instruct     & 0.0154 & \textcolor{High}{0.0543} & 0.0284 & 0.0072 & 0.0240 \\
        Qwen-2.5-14B-Instruct    & 0.0139 & \textcolor{High}{0.0573} & 0.0172 & 0.0000 & 0.0113 \\
        Qwen-2.5-32B-Instruct    & 0.0000 & 0.0000 & 0.0000 & \textcolor{High}{0.0073} & 0.0000 \\
        \textit{Mean ± Std}      
            & \textit{0.02 ± 0.02} 
            & \textit{0.05 ± 0.03} 
            & \textit{0.04 ± 0.03} 
            & \textit{0.03 ± 0.05} 
            & \textit{—} \\
        \midrule
        Qwen-2.5-coder-0.5B-Instruct  & 0.0195 & \textcolor{High}{0.0738} & 0.0480 & 0.0483 & 0.0615 \\
        Qwen-2.5-coder-1.5B-Instruct  & 0.0451 & 0.0465 & \textcolor{High}{0.0752} & 0.0284 & 0.0500 \\
        Qwen-2.5-coder-3B-Instruct    & 0.0000 & 0.0380 & \textcolor{High}{0.0406} & 0.0079 & 0.0328 \\
        Qwen-2.5-coder-7B-Instruct    & 0.0000 & 0.0382 & \textcolor{High}{0.0622} & 0.0189 & 0.0373 \\
        Qwen-2.5-coder-14B-Instruct   & 0.0189 & \textcolor{High}{0.0828} & 0.0372 & 0.0000 & 0.0313 \\
        Qwen-2.5-coder-32B-Instruct   & 0.0000 & \textcolor{High}{0.0786} & 0.0433 & 0.0356 & 0.0354 \\
        \textit{Mean ± Std}           
            & \textit{0.01 ± 0.02} 
            & \textit{0.06 ± 0.02} 
            & \textit{0.05 ± 0.01} 
            & \textit{0.02 ± 0.02}
            & \textit{—} \\
        \bottomrule
        \end{tabular}
        }
        \caption{HMI across logic change types for Qwen-2.5 Instruct and Coder variants on MBPP+. HMI highlighted in \textcolor{High}{red} denotes the highest HMI per row.}
        \label{tab:qwen_mri_mbpp_by_type}
\end{table*}
\definecolor{High}{RGB}{220, 20, 60}   
\definecolor{Mid}{RGB}{255,140,0}     

\begin{table*}[t]
    \centering
        \centering
        \renewcommand{\arraystretch}{1.22}
        \small
        \resizebox{\textwidth}{!}{
        \begin{tabular}{lrrrrrr}
         & \multicolumn{5}{c}{\textbf{HMI on BigCodeBench}} \\
        \cmidrule(lr){2-7}
        \textbf{Model} 
        & Component Swap 
        & Constraint Change 
        & Data Transformation 
        & Logic Reversal 
        & Workflow Modification
        & \textbf{All} \\
        \midrule
        Qwen-2.5-0.5B-Instruct   & N/A & N/A & N/A & N/A & N/A & N/A \\
        Qwen-2.5-1.5B-Instruct   & 0.1316 & 0.1551 & 0.1539 & \textcolor{High}{0.1698} & 0.0749 & 0.1444 \\
        Qwen-2.5-3B-Instruct     & 0.1334 & 0.1442 & 0.1399 & \textcolor{High}{0.1698} & 0.1558 & 0.1403 \\
        Qwen-2.5-7B-Instruct     & 0.1178 & 0.1355 & 0.1368 & \textcolor{High}{0.2118} & 0.1035 & 0.1296 \\
        Qwen-2.5-14B-Instruct    & 0.0685 & 0.0775 & \textcolor{High}{0.0998} & 0.0932 & 0.0850 & 0.0841 \\
        Qwen-2.5-32B-Instruct    & 0.0960 & 0.1130 & 0.1217 & 0.1268 & \textcolor{High}{0.1335} & 0.1143 \\
        \textit{Mean ± Std}      
            & \textit{0.12 ± 0.03} & \textit{0.14 ± 0.04} 
            & \textit{0.14 ± 0.03} & \textit{0.16 ± 0.05} 
            & \textit{0.12 ± 0.04} & \textit{—} \\
        \midrule
        Qwen-2.5-coder-0.5B-Instruct  & 0.1628 & 0.1302 & \textcolor{High}{0.2060} & 0.0787 & 0.1203 & 0.1778 \\
        Qwen-2.5-coder-1.5B-Instruct  & 0.1302 & \textcolor{High}{0.2207} & 0.1650 & 0.2135 & 0.1742 & 0.1630 \\
        Qwen-2.5-coder-3B-Instruct    & 0.1385 & 0.1813 & 0.1848 & \textcolor{High}{0.2111} & 0.0841 & 0.1630 \\
        Qwen-2.5-coder-7B-Instruct    & 0.1321 & 0.1440 & 0.1581 & \textcolor{High}{0.1947} & 0.1590 & 0.1477 \\
        Qwen-2.5-coder-14B-Instruct   & 0.1192 & 0.1110 & 0.1336 & 0.0956 & \textcolor{High}{0.1353} & 0.1237 \\
        Qwen-2.5-coder-32B-Instruct   & 0.1029 & 0.1242 & 0.1276 & 0.1191 & \textcolor{High}{0.1279} & 0.1178 \\
        \textit{Mean ± Std}           
            & \textit{0.13 ± 0.02} & \textit{0.15 ± 0.04} 
            & \textit{0.16 ± 0.03} & \textit{0.15 ± 0.06} 
            & \textit{0.13 ± 0.03} & \textit{—} \\
        \bottomrule
        \end{tabular}
        }
        \caption{HMI across logic change types for Qwen-2.5 Instruct and coder variants on BigCodeBench. HMI highlighted in \textcolor{High}{red} denotes the highest HMI per row. HMI is reported only for model--benchmark pairs satisfying the competence threshold in Section 3.1.1.}
        \label{tab:qwen_mri_bcb_by_type}
\end{table*}

\begin{table*}
    \centering
    \begin{minipage}{\textwidth}
        \centering
        \renewcommand{\arraystretch}{1.22}
        \small
        \resizebox{\textwidth}{!}{
               \begin{tabular}{l c ccc c ccc}
            \toprule
            \multirow{2}{*}{Model}
            & \multicolumn{4}{c}{\textbf{MBPP+}}
            & \multicolumn{4}{c}{\textbf{BigCodeBench}} \\
            \cmidrule(lr){2-5} \cmidrule(lr){6-9}
            
            & $\text{Acc}(\mathcal{T}_{\text{ori}})$ ($\uparrow$)
            & $\text{Sim}(\mathcal{T}_{\text{rew}})$ ($\downarrow$)
            & $\mathrm{RAD}_{\text{rew}}$ ($\downarrow$)
            & $\textbf{HMI}$ ($\downarrow$)
            & $\text{Acc}(\mathcal{T}_{\text{ori}})$ ($\uparrow$)
            & $\text{Sim}(\mathcal{T}_{\text{rew}})$ ($\downarrow$)
            & $\mathrm{RAD}_{\text{rew}}$ ($\downarrow$)
            & $\textbf{HMI}$ ($\downarrow$)\\

            \midrule
            Llama-3.1-8B-Instruct        & 0.5529   & 0.1486   & 0.0133   & 0.0020 & 0.3079   & 0.2132   & 0.4444   & 0.0947 \\
            Llama-3.1-70B-Instruct       & 0.6984   & 0.1518   & 0.0000   & 0.0000 & 0.4175   & 0.2404   & 0.3676   & 0.0884 \\
            \midrule
            Llama-4-Scout-17B-Instruct (16E)                & 0.6614   & 0.1446   & 0.0160   & 0.0023 & 0.4061   & 0.2343   & 0.3909   & 0.0916 \\
            Llama-4-Maverick-17B-Instruct (128E)             & 0.7751   & 0.2669   & 0.0307   & 0.0082 & 0.4860   & 0.2357   & 0.3953   & 0.0932 \\
            \bottomrule
        \end{tabular}

        }
        \caption{HMI for Llama-3.1 and Llama-4 instruct series. HMI persists in harder tasks (\textsc{BigCodeBench}), as $\mathrm{RAD}_{\text{rew}}$ stays high even as $\text{Sim}(\mathcal{T}_{\text{rew}})$ is comparable.}
        \label{mri_results_llama}
    \end{minipage}
\end{table*}

\begin{figure*}
    \centering
    \includegraphics[width=\linewidth]{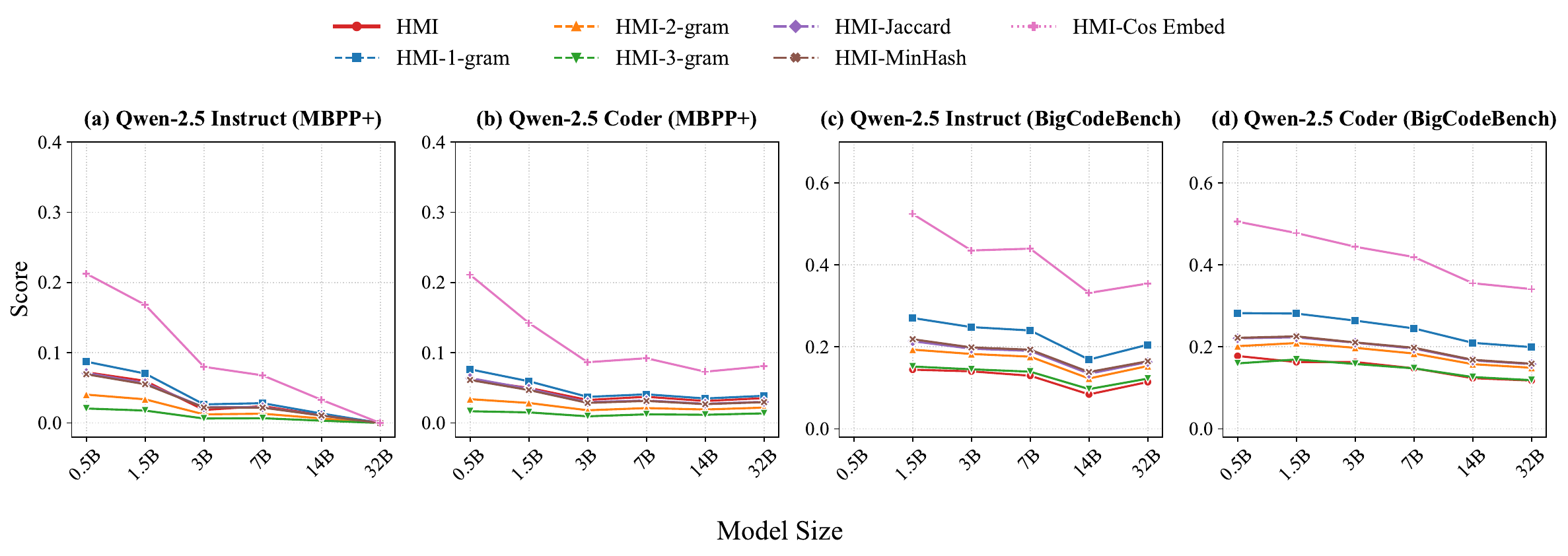}
    \caption{Scaling trends in HMI across Qwen-2.5 with different implementation choices.}
    \label{fig:ablation}
\end{figure*}

\begin{figure*}[t]
    \centering
    \includegraphics[width=\linewidth]{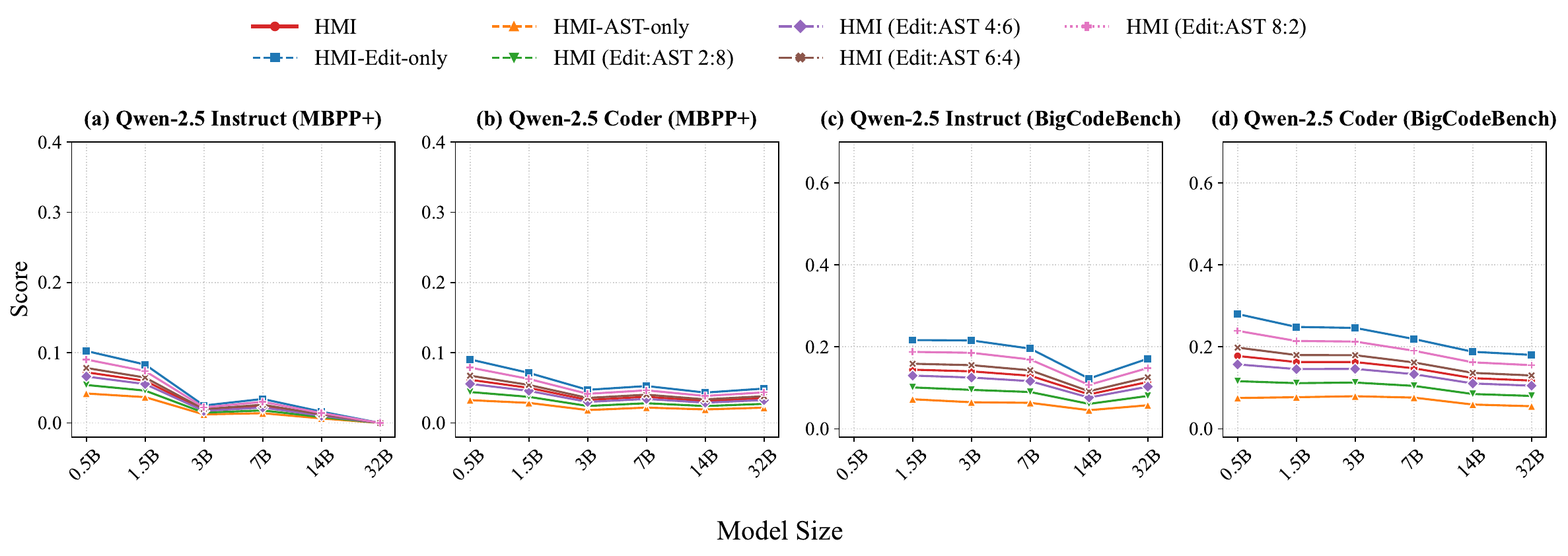}
    \caption{Scaling trends in HMI across Qwen-2.5 family with different weighting between AST-based and edit-based similarity.}
    \label{fig:ablation_weighting}
    \vspace{-8pt}
\end{figure*}

\label{sec:memorization_analysis on instruct models}

\paragraph{HMI does not monotonically increase with model scale.}
\autoref{fig:mem-qwen-series} shows the results for the Qwen-2.5 family on HMI and iHMI. Across both Qwen-2.5 Instruct and Qwen-2.5 Coder Instruct models, scaling is generally associated with lower $\mathrm{RAD}_{\mathrm{rew}}$ and hence lower HMI. This suggests that larger models are often better able to adapt to semantic rewrites, even when their outputs remain similar to the original reference solutions. The gains are most pronounced on simpler tasks (\textsc{MBPP+}) and are less consistent on harder tasks (\textsc{BigCodeBench}); on \textsc{BigCodeBench}, non-zero HMI and iHMI are primarily explained by persistently high $\mathrm{RAD}_{\mathrm{rew}}$ together with relatively stable $\mathrm{Sim}(\mathcal{T}_{\mathrm{rew}})$. Notably, as shown by the $\mathrm{Sim}(\mathcal{T}_{\mathrm{rew}})$ curves in~\autoref{fig:mem-qwen-series}(a) and (b), similarity to the original reference solution often remains high or even increases with model size. A similarity-only analysis would therefore suggest greater reuse of the original solution pattern. HMI gives a different view: when high similarity is not accompanied by a large performance drop, it is less indicative of the targeted high-similarity failure pattern and more consistent with benign reuse of common code structure.

\paragraph{HMI declines rapidly on simpler tasks but remains substantial on more difficult ones.}
On the introductory-level tasks in \textsc{MBPP+}, HMI decreases notably as models scale up. For instance, Qwen-2.5-Instruct’s HMI falls from Qwen-2.5-0.5B-Instruct to effectively zero at Qwen-2.5-32B-Instruct. Conversely, on the more challenging \textsc{BigCodeBench}, HMI values remain significant at Qwen-2.5-32B-Instruct. This discrepancy shows that while larger models better capture underlying semantics changes, non-zero HMI persists on challenging tasks, indicating continued brittleness under semantic rewrites.

\paragraph{Coder models increase solution similarity without substantially increasing HMI.}
Coder models yield higher $\mathrm{Sim}(\mathcal{T}_{\mathrm{rew}})$ than their instruction-only counterparts. For instance, on \textsc{BigCodeBench}, the Qwen-2.5-Coder series scores $\mathbf{0.3237} \pm 0.0086$ compared with $\mathbf{0.2670} \pm 0.0268$ for the instruction-only variant (mean $\pm$ SD across six model sizes). However, $\mathrm{RAD}_{\mathrm{rew}}$ remains comparable across these variants, translating to only a slight increase in HMI ($\mathbf{0.1367} \pm 0.0224$ vs.\ $\mathbf{0.1142} \pm 0.0247$). This pattern suggests that code-focused pretraining encourages reuse of familiar code structures, but this increased similarity does not necessarily translate into substantially higher HMI or larger rewrite-induced performance degradation.

\paragraph{Prompt-source control.}
To separate semantic changes from prompt-source effects, we additionally ask the same LLM to reword each original benchmark prompt while keeping the task semantics, reference solution, function signature, and unit tests unchanged. On MBPP+, semantics-preserving rewording yields a mean RAD of 0.000, compared with 0.118 for semantic rewrites. On BigCodeBench, among models satisfying our original-task competence threshold, the corresponding mean RAD values are 0.054 and 0.459, respectively. Thus, prompt rewording alone produces substantially smaller degradation, suggesting that prompt-source asymmetry or wording changes are unlikely to be the primary explanation for the observed performance drop. Full per-model results are provided in~\autoref{app:prompt_rewording}. This control does not rule out distribution shift or more general failures to adapt to modified task semantics.

\subsection{HMI concentrates in specific logic-edit types.}
To better understand which semantic changes are associated with higher HMI and rewrite brittleness, we annotate each rewritten pair with a fine-grained logic-change category, shown in~\autoref{tab:logic-change-types-merged}. To avoid unstable estimates, we exclude logic-change categories with fewer than 50 samples in a given dataset. Detailed definitions can be found in~\autoref{appendix:logic-change-types}. \autoref{tab:qwen_mri_mbpp_by_type} and~\autoref{tab:qwen_mri_bcb_by_type} show HMI by logic-change type. On \textsc{MBPP+} (\autoref{tab:qwen_mri_mbpp_by_type}), \textbf{Constraint Change} and \textbf{Data Transformation} consistently exhibit the highest per-type HMI across both Instruct and Coder series, whereas \textbf{Component Swap} has much smaller HMI and often approaches zero for larger models. On \textsc{BigCodeBench} (\autoref{tab:qwen_mri_bcb_by_type}), all five major types show substantial and relatively flat HMI across scales, with \textbf{Data Transformation}, \textbf{Logic Reversal}, and \textbf{Workflow Modification} frequently attaining the highest per-type HMI.

\subsection{Ablation Study}
We perform ablation studies to test whether our findings are robust across model families and alternative implementations of the similarity component in HMI.

\paragraph{Across Llama families.}
We evaluate Llama models in~\autoref{mri_results_llama}. The Llama-3.1 models show lower HMI as scale increases, consistent with the Qwen-family trend. The Llama~4 variants\footnote{The two Llama-4 variants we evaluate are MoE models with similar per-token activated compute; their difference is mainly \emph{capacity} (number of experts) rather than dense compute scaling.} show comparable HMI on both datasets. These results suggest that the main scaling pattern is not specific to the Qwen family, although the magnitude of HMI varies across model families and architectures.

\paragraph{Varying similarity functions.}
\autoref{fig:ablation} evaluates alternative similarity functions, including 1--3-gram overlap, Jaccard similarity, MinHash, and cosine similarity over code embeddings on both \textsc{MBPP+} and \textsc{BigCodeBench}. HMI trends are stable across these alternatives: Jaccard, MinHash, and 1--3-gram variants closely track our original HMI, while cosine-embedding similarity differs more in absolute scale but preserves the relative model trends. This indicates that our main conclusions are not an artifact of a single similarity implementation.

\paragraph{Varying the weighting between AST-based and edit-based similarity.}
\autoref{fig:ablation_weighting} evaluates different weightings between AST and edit-distance-based similarity. Across these settings, HMI under our default 50/50 weighting is highly correlated with HMI under alternative weightings. This suggests that our findings are not sensitive to the weighting of structural and edit-based similarity.

\begin{figure}
    \centering
    \includegraphics[width=1\linewidth, ]{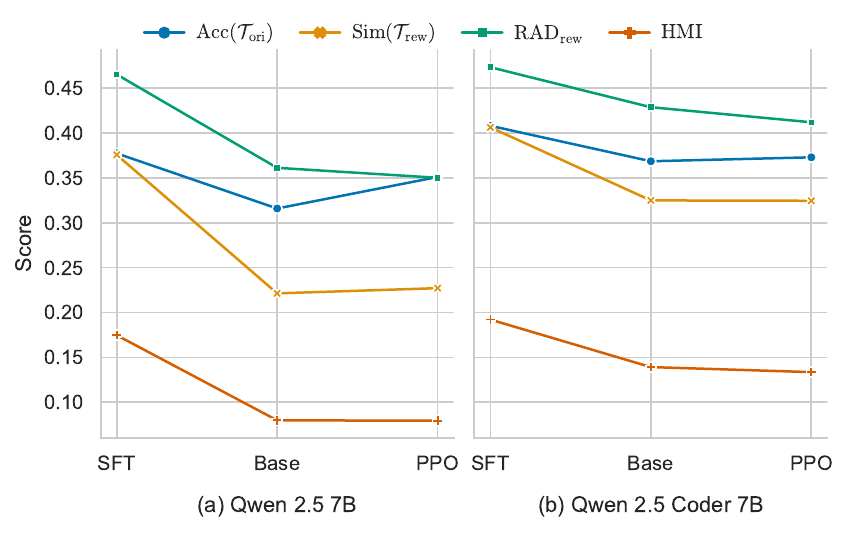}
    \caption{Results from fine-tuning models on \textsc{BigCodeBench}. Checkpoints selected for SFT and PPO follow the rules in Section~\ref{finetuning_methods}.}
    \label{fig:finetune bcb}
    \vspace{-8pt}
\end{figure}

\subsection{Impact of Fine-Tuning Strategies on HMI}
\paragraph{SFT improves original-task accuracy but increases HMI; PPO yields a better accuracy--HMI trade-off.}
\autoref{fig:finetune bcb} shows differences across post-training strategies on Qwen-2.5-7B and Qwen-2.5-Coder-7B on \textsc{BigCodeBench}. SFT consistently improves original-task accuracy. However, these gains are accompanied by substantial increases in HMI. These trends indicate that, under our semantic-rewriting diagnostic, SFT can improve original-task accuracy while increasing brittleness to rewritten semantics. Because SFT is trained on the training split and HMI is evaluated on held-out test tasks, this increase should not be interpreted as memorization of the evaluated task--solution pairs. Rather, it indicates increased reliance on learned solution patterns or reduced adaptation to rewritten semantics on held-out tasks. In contrast, in our evaluated settings, PPO maintains competitive original-task accuracy while keeping HMI substantially lower than SFT. Overall, PPO yields a more favorable accuracy--HMI trade-off, keeping HMI near or below base levels while still maintaining competitive original-task accuracy. Table~\ref{tab:sft_mri_trends} further shows that later SFT checkpoints can yield small additional accuracy gains but disproportionately larger HMI increases, suggesting that HMI can serve as a lightweight checkpoint-selection diagnostic when validation accuracy alone is insufficient.

\paragraph{Cross-family validation.} To examine whether the observed post-training pattern is specific to Qwen-2.5, we additionally conduct the same SFT/PPO comparison on Llama-3.1-8B using the same data split, training protocol, and checkpoint-selection criteria. The qualitative trend is consistent with the Qwen results: SFT increases original-task accuracy from 0.2588 to 0.2763 while increasing HMI from 0.0717 to 0.1816, whereas PPO achieves an original-task accuracy of 0.2675 with an HMI of 0.0801. These results provide an additional model-family case study supporting the observed accuracy--HMI trade-off, while we restrict this conclusion to the evaluated models and training settings. Full results are reported in~\autoref{app:llama_posttraining}.

\begin{table}
    \centering
    \small
    \resizebox{\linewidth}{!}{%
    \begin{tabular}{l l c c}
        \toprule
        Model & Stage (loss) & $\text{Acc}(\mathcal{T}_{\text{ori}})$ ($\uparrow$) & HMI \\
        \midrule
        \multirow{3}{*}{Qwen-2.5-7B}
        & Early (0.324)& 35.96\% & 0.1540 \\
        & Best-Val (0.302)& 37.72\% & 0.1733 \\
        & Late (0.326)& 39.47\% & 0.1795 \\
        \midrule
        \multirow{3}{*}{Coder-7B}
        & Early (0.325)& 35.96\% & 0.1537 \\
        & Best-Val (0.292)& 40.79\% & 0.1864 \\
        & Late (0.318)& 42.54\% & 0.2216 \\
        \bottomrule
    \end{tabular}
    }
    \caption{Changes in original-task accuracy and HMI during SFT training on \textsc{BigCodeBench}. Accuracy increases steadily from Early to Best-Val to Late, but HMI grows more sharply}
    \label{tab:sft_mri_trends}
    \vspace{-8pt}
    
\end{table}


\section{Conclusion and Future Work}

\label{sec:conclusion}


In this paper, we study the behavioral failure mode we operationalize as harmful memorization: failure under high similarity after task semantics change. We introduce code rewriting to construct superficially similar but semantically different task variants, and HMI to summarize whether models fail while retaining similarity to the original solution. Across MBPP+ and BigCodeBench, we find that HMI does not monotonically increase with scale, concentrates in specific logic-edit types, and changes substantially under post-training: in our evaluated settings, SFT can increase HMI despite improving original-task accuracy, while PPO yields a better accuracy--HMI trade-off. For practical use, HMI should be interpreted jointly with original-task accuracy: high accuracy with low HMI indicates both baseline competence and adaptation to semantic rewrites, whereas high accuracy with high HMI flags rewrite brittleness; low HMI with low baseline accuracy is inconclusive. Future work should extend this diagnostic beyond function-level tasks to repository-level settings and multi-file semantic changes.

\section{Limitations}
\label{sec:limitations}

Our diagnostic focuses on standalone function-level coding tasks and does not directly evaluate repository-level software engineering settings such as multi-file refactoring, long-horizon dependency reasoning, or SWE-Bench-style issue resolution. Our semantic rewrites are localized logic changes, which support controlled attribution but may not capture broader requirement changes in real-world codebases. HMI is an operational behavioral diagnostic rather than direct evidence of training-set memorization. HMI should also be interpreted together with original-task accuracy: low HMI does not necessarily imply strong coding ability when the model has low baseline accuracy. Our prompt-rewording control provides evidence against prompt-wording sensitivity or prompt-source differences as the primary explanation for the observed degradation, but it does not eliminate alternative explanations such as distribution shift or a more general failure to adapt to modified task semantics. Finally, our rewriting pipeline uses an LLM judge for dataset validation, although candidate model scoring itself relies on executable tests and deterministic similarity measures.

\section{Acknowledgements}
\paragraph{Use of AI Assistance}
We used GPT-5 for generating candidate rewritten tasks and for validating prompt--solution consistency during dataset construction. Candidate model evaluation does not rely on GPT-5; it uses executable unit tests and deterministic similarity metrics. We also used GPT-5 for language polishing. All research ideas, experimental design, analysis, and conclusions were developed and verified by the authors.

\section{Ethics Statement}

Our \emph{code rewriting} pipeline is guided by ethical principles to ensure responsible outcomes. 

(1) Data:  Our dataset is constructed from the MBPP+ and BIGCODEBENCH datasets. We aim to exclude harmful or offensive content from the rewritten datasets to mitigate potential risks.

(2) Responsible Usage and License: The use of the \emph{code rewriting} dataset is intended solely for evaluating memorization in LLM code generation tasks. We encourage the responsible use of these datasets for educational and scientific purposes, while strongly discouraging any harmful or malicious activities.



\bibliography{custom}

\appendix
\section*{Appendix}

\renewcommand{\sectionautorefname}{Appendix}
\renewcommand{\subsectionautorefname}{Appendix}

\section{Prompts for Task Generation}
\label{appendix:prompts}
\label{appendix:prompts}

We provide the full instruction prompts used to generate \emph{code rewriting} variant with GPT-5. The system and user messages are shown as passed to the API.
\definecolor{airforceblue}{rgb}{0.36, 0.54, 0.66}
\newtcolorbox[list inside=prompt,auto counter,number within=section]{prompt}[1][]{
    colbacktitle=airforceblue,
    colframe=airforceblue,
    fontupper=\footnotesize,
    boxsep=5pt,
    left=0pt,
    right=0pt,
    top=0pt,
    bottom=0pt,
    boxrule=1pt,
    enhanced, 
    breakable,
    skin first=enhanced,
    skin middle=enhanced,
    skin last=enhanced,
    #1,
}
\begin{prompt}[title={System Prompt},label=prompt:evolve:system]
\promptsubsection{System}
You are an experienced python programmer. Your goal is to transforms a given 'coding task prompt' into a new version. Follow the instructions carefully to transform the prompt.
\end{prompt}

\begin{prompt}[title={Code-Rewriting Evolution User Prompt},label=prompt:evolve:rewrite]
\promptsubsection{User}
\begin{PromptVerbatim}
Given a coding task description (#The Given Prompt#) and its canonical solution (#Code#),
    perform the following steps:
1. Modify the canonical solution to create #New Code# by altering only ***ONE*** core 
    logic or structure. Do not add additional 'if statements' to the code. Avoid 
    superficial changes like variable renaming. Ensure the modified code has different 
    semantics in a way that ***expected difficulty equivalent to the original problem***. 
    Write a #New Entry Point# to the updated code. This function name must be very 
    similar or the same as the old entry point, and reflect the modified code's logic 
    changes if using #Old Entry Point# could mislead the programmer on the 
    #Rewritten Prompt#.
2. Update #The Given Prompt# to create #Rewritten Prompt#. The new prompt must:"
    - Match the original's ***input signature*** exactly, but the output 
        format could be different a little bit.
    - Reflect the modified code's logic changes explicitly.
        Retain the original phrasing structure and ***avoid unnecessary rephrasing*** 
        in a way that the #Rewritten Prompt# syntactically very similar 
        to the #The Given Prompt#.
3. If any mismatch arises between new code and new prompt, revise either one 
    (without adding more changes) so all constraints in Steps 1-2 are simultaneously 
    satisfied.


Format your response exactly as:

New Code:
[code]

Explanation:
[logic changes]

Rewritten Prompt:
[updated description]

Old Entry Point:
[original function name]

New Entry Point:
[updated function name]

\end{PromptVerbatim}
\end{prompt}

\subsection{LLM Judge}

\begin{prompt}[title={System Prompt},label=prompt:evolve:system]
\promptsubsection{System}
You are an expert code reviewer. Your task is to evaluate whether an evolved coding task maintains appropriate quality standards in terms of prompt-code alignment and difficulty equivalence.
\end{prompt}

\begin{prompt}[title={Code-Rewriting Evolution LLM Judging Prompt},label=prompt:evolve:rewrite]
\promptsubsection{User}
Please evaluate the quality of this evolved coding task by analyzing two key aspects:
\begin{PromptVerbatim}
**Original Task:**
Prompt: {original_prompt} 
Code: {original_code}
**Evolved Task:**
Prompt: {rewritten_prompt}
Code: {rewritten_code}

**Evaluation Criteria:** 
1. **Prompt-Code Alignment**: Does the new prompt accurately describe what the new code 
    does?
   - Are the input/output specifications consistent?
   - Does the prompt clearly communicate the expected behavior?
   - Are there any ambiguities or mismatches?

2. **Difficulty Equivalence**: Is the evolved task of similar difficulty to the original?
   - Does it require similar algorithmic thinking?
   - Is the complexity level maintained (not significantly easier or harder)?
   - Does it test similar programming concepts and skills?

**Response Format:** 
Provide your evaluation in the following format:

Alignment Score: [1-5, where 5 = perfect alignment, 1 = major misalignment]
Alignment Reasoning: [Brief explanation of why the prompt and code align or don't align]

Difficulty Score: [1-5, where 5 = equivalent difficulty, 3 = acceptable variation, 1 = 
    significantly different]
Difficulty Reasoning: [Brief explanation of difficulty comparison]

Overall Recommendation: [ACCEPT/REJECT]
Overall Reasoning: [Brief summary of your decision]
Please be thorough but concise in your evaluation.
\end{PromptVerbatim}
\end{prompt}

\section{Code Rewriting Ground Truth Solution Validation}
\label{Code Rewriting Solution Validation}
Additionally, we ensured the validity of test cases for all rewritten tasks across both datasets, and validate each rewritten solution by making it pass its corresponding rewritten unit test. For MBPP+, we reuse the official test case inputs and generate the expected outputs using the rewritten ground-truth solutions, ensuring direct comparability. For BigCodeBench, we adopt the procedure outlined in \cite{zhuo2024bigcodebench}, constructing test cases for each rewritten task based on their guidelines to guarantee consistency and correctness. We installed all packages required by both dataset for assessing function correctness.

\section{Human Validation of Rewritten Tasks}
\label{appendix:validation}
\begin{table}[!ht]
\centering
\small
\caption{Human validation of rewritten tasks.}
\label{tab:data_validation}
\setlength{\tabcolsep}{4pt}
\begin{tabularx}{\columnwidth}{
    @{}>{\raggedright\arraybackslash}Xcc@{}
}
\toprule
Metric & MBPP+ & BigCodeBench \\
\midrule
\# Pairs
& 378
& 1140 \\

Alignment score
& $4.95 \pm 0.26$
& $4.81 \pm 0.31$ \\

Difficulty equiv. score
& $4.89 \pm 0.39$
& $4.73 \pm 0.56$ \\

Easier / Similar / Harder
& 5\% / 91\% / 4\%
& 10\% / 79\% / 11\% \\

First-pass acceptance
& 97.6\%
& 94.9\% \\
\bottomrule
\end{tabularx}
\end{table}

Annotators used the same criteria as the LLM judge to rate prompt--solution alignment, difficulty equivalence, and overall acceptance. Pairs receiving a majority rejection or no majority decision were returned to the rewriting pipeline for regeneration and re-evaluation. Although exact difficulty equality cannot be guaranteed for every individual pair, the aggregate results provide no evidence of a systematic shift toward harder rewritten tasks.


\section{Prompt-Rewording Control}
\label{app:prompt_rewording}
LLMs can be sensitive to prompt wording, and our semantic-rewriting pipeline generates rewritten prompts with an LLM. We therefore conduct an additional control to examine whether the observed performance degradation can be primarily attributed to prompt rewording rather than to the semantic changes introduced by our rewriting procedure.

For each original benchmark task, we ask the same LLM to reword the original prompt while preserving the task semantics, reference solution, function signature, and unit tests. We denote these semantics-preserving reworded tasks as $T_{\mathrm{para}}$. We then compare model performance on three versions of each task: the original benchmark task $T_{\mathrm{ori}}$, the semantics-altering rewritten task $T_{\mathrm{rew}}$, and the semantics-preserving reworded task $T_{\mathrm{para}}$.

For this control, we compute the relative accuracy drop analogously to Eq.~\ref{eq:rad_rew}:
\begin{equation}
\mathrm{RAD}_{\mathrm{para}}
=
\max\left(
0,
1 -
\frac{\mathrm{Acc}(T_{\mathrm{para}})}
{\mathrm{Acc}(T_{\mathrm{ori}})}
\right).
\end{equation}

If the degradation observed on $T_{\mathrm{rew}}$ were mainly caused by LLM-generated wording or prompt-source differences, we would expect $\mathrm{RAD}_{\mathrm{para}}$ to be comparable to $\mathrm{RAD}_{\mathrm{rew}}$.

\begin{table*}[t]
\centering
\small
\begin{tabular}{lccccc}
\toprule
Model &
$\mathrm{Acc}_{\mathrm{ori}}$ &
$\mathrm{Acc}_{\mathrm{rew}}$ &
$\mathrm{Acc}_{\mathrm{para}}$ &
$\mathrm{RAD}_{\mathrm{rew}}$ &
$\mathrm{RAD}_{\mathrm{para}}$ \\
\midrule
Qwen-2.5-0.5B-Instruct  & 0.4021 & 0.2937 & 0.4180 & 0.2697 & 0.0000 \\
Qwen-2.5-1.5B-Instruct  & 0.5767 & 0.4550 & 0.5952 & 0.2110 & 0.0000 \\
Qwen-2.5-3B-Instruct    & 0.6243 & 0.5608 & 0.6323 & 0.1017 & 0.0000 \\
Qwen-2.5-7B-Instruct    & 0.6852 & 0.6270 & 0.6905 & 0.0849 & 0.0000 \\
Qwen-2.5-14B-Instruct   & 0.7037 & 0.6746 & 0.7116 & 0.0414 & 0.0000 \\
Qwen-2.5-32B-Instruct   & 0.7513 & 0.7725 & 0.7593 & 0.0000 & 0.0000 \\
\midrule
Avg. $\pm$ Std. &
0.624 $\pm$ 0.125 &
0.564 $\pm$ 0.170 &
0.634 $\pm$ 0.121 &
0.118 $\pm$ 0.103 &
0.000 $\pm$ 0.000 \\
\bottomrule
\end{tabular}
\caption{Prompt-rewording control on MBPP+. 
$T_{\mathrm{rew}}$ changes task semantics, whereas
$T_{\mathrm{para}}$ preserves the original task semantics and only rewords the prompt.}
\label{tab:prompt_control_mbpp}
\end{table*}

\begin{table*}[t]
\centering
\small
\begin{tabular}{lccccc}
\toprule
Model &
$\mathrm{Acc}_{\mathrm{ori}}$ &
$\mathrm{Acc}_{\mathrm{rew}}$ &
$\mathrm{Acc}_{\mathrm{para}}$ &
$\mathrm{RAD}_{\mathrm{rew}}$ &
$\mathrm{RAD}_{\mathrm{para}}$ \\
\midrule
Qwen-2.5-0.5B-Instruct  & 0.0947 & 0.0325 & 0.0956 & N/A & N/A \\
Qwen-2.5-1.5B-Instruct  & 0.2281 & 0.0956 & 0.2360 & 0.5808 & 0.0000 \\
Qwen-2.5-3B-Instruct    & 0.3132 & 0.1649 & 0.2991 & 0.4734 & 0.0448 \\
Qwen-2.5-7B-Instruct    & 0.3798 & 0.1965 & 0.3465 & 0.4827 & 0.0878 \\
Qwen-2.5-14B-Instruct   & 0.3895 & 0.2456 & 0.3658 & 0.3694 & 0.0608 \\
Qwen-2.5-32B-Instruct   & 0.4404 & 0.2702 & 0.4070 & 0.3865 & 0.0757 \\
\midrule
Avg. $\pm$ Std. &
-- &
-- &
-- &
0.459 $\pm$ 0.076 &
0.054 $\pm$ 0.031 \\
\bottomrule
\end{tabular}
\caption{Prompt-rewording control on BigCodeBench. 
RAD values are reported only for models satisfying the original-task competence threshold defined in Section~3.1.1. 
$T_{\mathrm{rew}}$ changes task semantics, whereas
$T_{\mathrm{para}}$ preserves the original task semantics and only rewords the prompt.}
\label{tab:prompt_control_bcb}
\end{table*}

Across the evaluated Qwen-2.5 Instruct models (\autoref{tab:prompt_control_mbpp} and~\autoref{tab:prompt_control_bcb}), semantics-preserving prompt rewording causes substantially smaller degradation than semantics-altering rewriting. On MBPP+, the mean $\mathrm{RAD}_{\mathrm{para}}$ is 0.000, compared with 0.118 for $\mathrm{RAD}_{\mathrm{rew}}$. On BigCodeBench, among models satisfying our original-task competence threshold, the corresponding mean RAD values are 0.054 and 0.459, respectively. These results suggest that LLM-generated wording or prompt-source differences alone are unlikely to be the primary explanation for the observed performance degradation. However, this control does not rule out other explanations, such as distribution shift or a more general failure to adapt to modified task semantics.

\section{Additional Post-Training Results on Llama-3.1-8B}
\label{app:llama_posttraining}
To further examine the generality of the post-training trends reported in Section~5.4, we repeat the SFT and PPO experiments using Llama-3.1-8B~(\autoref{tab:llama_posttraining}). We use the same data split, training protocol, and checkpoint-selection criteria as in the corresponding Qwen experiments. SFT and PPO are independently initialized from the same base Llama-3.1-8B model.

\begin{table}[!ht]
\centering
\small
\setlength{\tabcolsep}{4pt}
\begin{tabular}{lcccc}
\toprule
Model &
$\mathrm{Acc}_{\mathrm{ori}}$ &
$\mathrm{Sim}$ &
$\mathrm{RAD}_{\mathrm{rew}}$ &
HMI \\
\midrule
Base & 0.2588 & 0.1922 & 0.3729 & 0.0717 \\
SFT  & 0.2763 & 0.4087 & 0.4444 & 0.1816 \\
PPO  & 0.2675 & 0.2223 & 0.3604 & 0.0801 \\
\bottomrule
\end{tabular}
\caption{Post-training results on Llama-3.1-8B. SFT and PPO are independently initialized from the same base model. Here, $\mathrm{Sim}$ denotes $\mathrm{Sim}(T_{\mathrm{rew}})$.}
\label{tab:llama_posttraining}
\end{table}

The Llama-3.1-8B results reproduce the qualitative pattern observed for Qwen-2.5. SFT yields the highest original-task accuracy but also substantially increases both similarity and RAD, resulting in a higher HMI. PPO improves original-task accuracy relative to the base model while keeping HMI much closer to the base level. This additional experiment suggests that the observed SFT/PPO contrast is not unique to the Qwen-2.5 models evaluated in our main experiments. However, these results remain limited to the model families, datasets, and training configurations considered in this study and should not be interpreted as establishing a general advantage of PPO over SFT.

\section{Training details regarding SFT/RL}
\label{appendix:finetuning_details}

\subsection{Fine-Tuning Details}
\label{app:finetune}

\paragraph{Framework and Compute.}
We adapted the \texttt{Verl} framework for supervised fine-tuning (SFT) and Proximal Policy Optimization (PPO), using its PyTorch Fully Sharded Data Parallel (FSDP) backend Experiments ran on a single machine equipped with four NVIDIA A100 GPUs, with each training or inference job using two GPUs (nnodes=1, n\_gpus\_per\_node=2). PPO rollouts used the \texttt{VLLM} backend; optimization used \texttt{AdamW}.

\begin{table}
  \centering
  \caption{Compute and framework configuration}
  \label{tab:compute}
  \begin{tabularx}{\linewidth}{lX}
    \toprule
    Item & Setting \\
    \midrule
    Framework & \texttt{Verl} (PyTorch FSDP backend) \\
    Nodes / GPUs & \texttt{nnodes=1}, \texttt{n\_gpus\_per\_node=2} \\
    PPO rollout backend & \texttt{VLLM} \\
    Optimizer & \texttt{AdamW} \\
    \bottomrule
  \end{tabularx}
\end{table}

\paragraph{Dataset and Prompting.}
Data followed \texttt{Verl}'s standard format and was exported as a \texttt{.parquet} file with a 4:1 train/test split. Each problem description served as the prompt; the corresponding code solution was the target response. 

\begin{table}
  \centering
  \caption{Dataset summary}
  \label{tab:data}
  \begin{tabularx}{\linewidth}{lX}
    \toprule
    Aspect & Details \\
    \midrule
    Format & \texttt{.parquet} (Verl standard) \\
    Split & 4:1 train:test \\
    Input (prompt) & Problem description \\
    Target (response) & Code solution \\
    Prompt template & See quoted block above \\
    \bottomrule
  \end{tabularx}
\end{table}

The completed template we fed into the LLM was:


\lstset{
  basicstyle=\ttfamily\small,
  breaklines=true,
  morestring=[b]",
  literate={"}{{\texttt{"}}}1
}
\begin{lstlisting}[language=Python]
instruction_prefix = "Please provide a self-contained Python script that solves the following problem in a markdown code block:" 

response_prefix = "Below is a Python script with a self-contained function that solves the problem and passes corresponding tests:" 

prompt_chat = [
    {"role": "user", "content": f"""\
{instruction_prefix}
```
{problem.strip()}
```
"""},
    {"role": "assistant", "content": f"""\
{response_prefix}
```python
"""}
]
\end{lstlisting}

The \textbf{problem} is the description originally from the dataset, and we called the \texttt{tokenizer.apply\_chat\_template} to the \textbf{prompt\_chat} to get the model response.

\subsubsection{Supervised Fine-Tuning (SFT)}
Default learning rate was \(1\times 10^{-5}\) for 20 epochs, with manual adjustments between \(5\times 10^{-6}\) and \(1\times 10^{-5}\) depending on model performance. We set \texttt{max\_prompt\_length} to 1024, \texttt{batch size} to 64, and \texttt{micro\_batch\_size\_per\_gpu} to 8. The selected checkpoint (named \textbf{model\_name-SFT}) was the one immediately prior to observed overfitting, allowing us to examine HMI before later-stage overfitting.

Moreover, we choose the checkpoint at epoch 20 (named \textbf{model\_name-SFT-overfit}) as a late-training checkpoint to examine how prolonged SFT affects HMI.

\begin{table}
  \centering
  \caption{SFT hyperparameters}
  \label{tab:sft}
  \begin{tabularx}{\linewidth}{lX}
    \toprule
    Parameter & Value \\
    \midrule
    Epochs & 20 \\
    Learning rate & Default \(1\times 10^{-5}\); tuned \(5\times 10^{-6}\)–\(1\times 10^{-5}\) \\
    max\_prompt\_length & 1024 \\
    Batch size & 64 \\
    micro\_batch\_size\_per\_gpu & 8 \\
    save\_freq & after\_each\_epoch \\
    Checkpoint selection & Epoch immediately prior to overfitting \\
    \bottomrule
  \end{tabularx}
\end{table}

\subsubsection{Proximal Policy Optimization (PPO)}
Actor, critic, and reference models used identical architectures over 20 epochs. The reward was binary: 1 if the generated response passed all test cases, else 0. We set \texttt{max\_prompt\_length} to 1024 and \texttt{max\_response\_length} to 512. Learning rates were \(1\times 10^{-5}\) for the critic and \(1\times 10^{-6}\) for the actor. We used \texttt{batch size} 64 with \texttt{micro\_batch\_size\_per\_gpu} 8, selecting the checkpoint with the highest validation reward (named \textbf{model\_name-PPO}) to get the best performance.

\begin{table}
  \centering
  \caption{PPO setup and hyperparameters}
  \label{tab:ppo}
  \begin{tabularx}{\linewidth}{lX}
    \toprule
    Parameter & Value \\
    \midrule
    Architectures & Actor/Critic/Reference identical \\
    Epochs & 20 \\
    Reward & Binary (1 if all tests pass; else 0) \\
    max\_prompt\_length & 1024 \\
    max\_response\_length & 512 \\
    Learning rate (critic) & \(1\times 10^{-5}\) \\
    Learning rate (actor) & \(1\times 10^{-6}\) \\
    Batch size & 64 \\
    micro\_batch\_size\_per\_gpu & 8 \\
    save\_freq & 5 \\
    Checkpoint selection & Highest reward on validset \\
    \bottomrule
  \end{tabularx}
\end{table}

\section{Evolved-Task Generation (GPT-5)}

\label{appendix_dataset}
\begin{itemize}[leftmargin=1.2em]
\item \textbf{API version}: \texttt{gpt-5-2025-08-07}.%
\item \textbf{Prompt template}: shown in~\autoref{appendix:prompts}.%
\item \textbf{Parameters}: temperature: default; top-p: default; max-tokens 1080.%
\item \textbf{Post-processing}: regex clean-up.%
\item \textbf{Budget}: the estimated cost for generating one round of each evolution type (code rewriting, mutation and paraphrase) for both MBPP+ and BigCodeBench is approximately 450 USD.
\end{itemize}

\section{Stability of RAD and HMI under low baseline accuracy}
\label{appendix:mri_stability}
The Relative Accuracy Drop (RAD) measures the \emph{fractional} decrease in Pass@1 between
the original and rewritten task sets. Although this captures the intended notion of
rewriting-sensitivity, it becomes unstable when the baseline accuracy
$\mathrm{Acc}(\mathcal{T}_{\mathrm{ori}})$ is extremely low: even a small absolute change can
inflate the relative ratio and consequently HMI (defined as the product of similarity and RAD).

To ensure that RAD and HMI reflect meaningful performance changes rather than numerical
instability, we compute these metrics only when two conditions hold.

\paragraph{(1) Sufficient number of correctly solved original tasks.}
For a benchmark with $N$ tasks and baseline accuracy $p = \mathrm{Acc}(\mathcal{T}_{\mathrm{ori}})$,
the model solves $k = Np$ tasks. We require
\[
k \ge M = 50,
\]
which keeps the variance of the empirical Pass@1 sufficiently small under a binomial model.
For instance, MBPP+ ($N=378$) requires $p >= 13.2\%$, resulting in a $95\%$ confidence
interval of roughly $\pm 3$–$4$ percentage points; BigCodeBench ($N=1{,}140$) requires
$p >=4.4\%$ with an even tighter interval. These uncertainties are far smaller than the
5–20 point drops typically observed in our memorization analyses.

\paragraph{(2) A minimum competence threshold on baseline accuracy.}
We additionally require
\[
\mathrm{Acc}(\mathcal{T}_{\mathrm{ori}}) \ge 10\%,
\]
which prevents the denominator in RAD from becoming so small that modest absolute changes
(e.g., a 5-point drop) produce disproportionally large relative decreases (e.g., $>50\%$).

\paragraph{Application of the threshold.}
When either condition is violated, we still report
$\mathrm{Acc}(\mathcal{T}_{\mathrm{ori}})$ and $\mathrm{Acc}(\mathcal{T}_{\mathrm{rew}})$ for
transparency, but we mark RAD and HMI as
\emph{``N/A (below competence threshold)''} and exclude them from cross-model comparisons.
All analyses and figures in the main paper adhere to this stability rule.

These criteria reduce numerical artifacts caused by very low baseline accuracy and make cross-model comparisons of RAD and HMI more stable.

\section{Analysis by Logic Change Type}
\definecolor{High}{RGB}{220, 20, 60}   
\definecolor{Mid}{RGB}{255,140,0}     

\label{appendix:logic-change-types}

\paragraph{Logic change types.}
To better understand \emph{which} semantic edits are associated with greater rewrite brittleness, we categorize each semantics-altering rewrite into one of the following nine logic change types:
\begin{itemize}[leftmargin=*]
    \item \textbf{Component Swap:} Swapping a function, class, or method while keeping the surrounding structure fixed (e.g., \texttt{sum}~$\to$~\texttt{max}, \texttt{np.std}~$\to$~\texttt{np.var}).
    \item \textbf{Data Transformation:} Transforming the data \emph{before} applying the main logic (e.g., \texttt{np.log(data)}, \texttt{data.T}, \texttt{data.pct\_change()}, tokenization or normalization steps).
    \item \textbf{Constraint Change:} Modifying rules, parameters, or logical conditions (e.g., loop stride in \texttt{range(..., 2)}, bounds in inequalities, regex patterns, or SQL \texttt{UNIQUE} constraints).
    \item \textbf{Logic Reversal:} Inverting the decision or ranking logic (e.g., \texttt{max}~$\to$~\texttt{min}, \texttt{sorted(...)}~$\to$~\texttt{sorted(..., reverse=True)}, swapping the roles of \texttt{x} and \texttt{y} in comparisons).
    \item \textbf{Operator Change:} Swapping a single arithmetic or logical operator (e.g., \texttt{\^{ } }~$\to$~\texttt{+}, \texttt{==}~$\to$~\texttt{!=}).
    \item \textbf{Operation Order Swap:} Reordering two or more operations with the same components (e.g., applying \texttt{blur} then \texttt{gray} vs.\ \texttt{gray} then \texttt{blur}).
    \item \textbf{Constant Swap:} Changing hard-coded constants or literal values (e.g., switching between \texttt{lowercase} and \texttt{UPPERCASE}, or adjusting thresholds).
    \item \textbf{Workflow Modification:} Structural changes to the overall pipeline, such as inserting or removing processing stages (e.g., adding an extra \texttt{zlib} compression step or altering nested outputs).
    \item \textbf{Other:} Any semantics-altering edit that does not clearly fall into the above categories.
\end{itemize}

\begin{table}[t]
\centering
\small
\resizebox{\columnwidth}{!}{%
\begin{tabular}{lrrrr}
\toprule
Logic Change Type & BigCodeBench \# & BigCodeBench \% & MBPP+ \# & MBPP+ \% \\
\midrule
Total                     & 1140 & 100.00\% & 378 & 100.00\% \\
Component Swap           & 393  & 34.47\%  & 52  & 13.76\% \\
Data Transformation      & 385  & 33.77\%  & 58  & 15.34\% \\
Constraint Change        & 185  & 16.23\%  & 86  & 22.75\% \\
Logic Reversal           & 50   & 4.39\%   & 68  & 17.99\% \\
Workflow Modification    & 50   & 4.39\%   & 10  & 2.65\%  \\
Constant Swap            & 37   & 3.25\%   & 36  & 9.52\%  \\
Operation Order Swap    & 20   & 1.75\%   & 12  & 3.17\%  \\
Operator Change          & 18   & 1.58\%   & 36  & 9.52\%  \\
Other                     & 2    & 0.18\%   & 20  & 5.29\%  \\
\bottomrule
\end{tabular}
}
\caption{Code Rewriting logic change type statistics for \texttt{bigcodebench} and \texttt{mbppplus}.}
\label{tab:logic-change-types-transposed}
\end{table}

\paragraph{Coverage across benchmarks.}
\autoref{tab:logic-change-types-transposed} reports the distribution of logic change types across \textsc{BigCodeBench} and \textsc{MBPP+}. On \textsc{BigCodeBench}, the changes are dominated by \textbf{Component Swap} (34.5\%) and \textbf{Data Transformation} (33.8\%), followed by \textbf{Constraint Change} (16.2\%), with the remaining types each accounting for less than 5\% of all instances. On \textsc{MBPP+}, the distribution is more balanced: \textbf{Constraint Change} (22.8\%), \textbf{Logic Reversal} (18.0\%), \textbf{Data Transformation} (15.3\%), \textbf{Component Swap} (13.8\%), and \textbf{Constant/Operator Swap} (9--10\%) all contribute substantially. To avoid unstable estimates, all per-type statistics in this appendix exclude logic-change categories with fewer than 50 samples in a given dataset.

\begin{table*}[t]
\centering
\small
\resizebox{\textwidth}{!}{%
\begin{tabular}{lrrrrrrrr}
\toprule
& \multicolumn{4}{c}{qwen-2.5-coder-0.5b-instruct} 
& \multicolumn{4}{c}{qwen-2.5-coder-14b-instruct} \\
\cmidrule(lr){2-5} \cmidrule(lr){6-9}
Logic Change Type 
& Total & $\text{Acc}(\mathcal{T}_{\text{ori}})$ ($\uparrow$) & $\text{Acc}(\mathcal{T}_{\text{rew}})$ ($\uparrow$) & $\text{RAD}$ ($\downarrow$) 
& Total & $\text{Acc}(\mathcal{T}_{\text{ori}})$ ($\uparrow$) & $\text{Acc}(\mathcal{T}_{\text{rew}})$ ($\uparrow$) & $\text{RAD}$ ($\downarrow$) \\
\midrule
Component Swap      & 52 & 48.1\% & 44.2\% & 8.0\%
                     & 52 & 75.0\% & 71.2\% & 5.1\% \\
Constraint Change   & 86 & 46.5\% & 30.2\% & \textcolor{High}{35.0\%}
                     & 86 & 73.3\% & 54.7\% & \textcolor{High}{25.4\%} \\
Data Transformation & 58 & 55.2\% & 41.4\% & 25.0\%
                     & 58 & 72.4\% & 62.1\% & 14.3\% \\
Logic Reversal      & 68 & 42.6\% & 35.3\% & 17.2\%
                     & 68 & 72.1\% & 76.5\% & 0.0\% \\
\midrule
\textbf{HMI (all nine types)} 
& \textbf{--} & \textbf{--} & \textbf{--} & \textbf{0.0615}
& \textbf{--} & \textbf{--} & \textbf{--} & \textbf{0.0313} \\
\midrule
& \multicolumn{4}{c}{qwen-2.5-0.5b-instruct} 
& \multicolumn{4}{c}{qwen-2.5-14b-instruct} \\
\cmidrule(lr){2-5} \cmidrule(lr){6-9}
Component Swap      & 52 & 44.2\% & 40.4\% & 8.7\%
                     & 52 & 75.0\% & 71.2\% & 5.1\% \\
Constraint Change   & 86 & 39.5\% & 24.4\% & \textcolor{High}{38.2\%}
                     & 86 & 64.0\% & 48.8\% & \textcolor{High}{23.6\%} \\
Data Transformation & 58 & 50.0\% & 32.8\% & 34.5\%
                     & 58 & 70.7\% & 65.5\% & 7.3\% \\
Logic Reversal      & 68 & 39.7\% & 26.5\% & 33.3\%
                     & 68 & 72.1\% & 80.9\% & 0.0\% \\
\midrule
\textbf{HMI (all nine types)} 
& \textbf{--} & \textbf{--} & \textbf{--} & \textbf{0.0722}
& \textbf{--} & \textbf{--} & \textbf{--} & \textbf{0.0113}\\
\bottomrule
\end{tabular}
}
\caption{Comparison of logic change type performance between four Qwen-2.5 variants on \texttt{MBPP+}. We omit logic change types with fewer than 50 samples to avoid statistical bias. RAD highlighted in \textcolor{High}{red} denotes the highest drop.}
\label{tab:qwen-mbppplus-logic-types}
\end{table*}

\paragraph{Per-type RAD on MBPP+.}
\autoref{tab:qwen-mbppplus-logic-types} shows accuracy on original tasks $\text{Acc}(\mathcal{T}_{\text{ori}})$, rewritten tasks $\text{Acc}(\mathcal{T}_{\text{rew}})$, and the resulting accuracy drop $\mathrm{RAD}$ (all for semantics-altering rewrites, i.e., $\mathrm{RAD}_{\text{rew}}$) for four Qwen-2.5 variants on \textsc{MBPP+}. Several patterns emerge:
\begin{itemize}[leftmargin=*]
    \item For both Instruct and Coder families, \textbf{Constraint Change} exhibits the largest $\mathrm{RAD}$ at both 0.5B and 14B, indicating that models are particularly brittle when loop bounds, thresholds, or other constraints are modified.
    \item \textbf{Data Transformation} also induces substantial drops (25--35\% at 0.5B; 7--14\% at 14B), showing the difficulty of tracking changes to data pre-processing or representation.
    \item \textbf{Component Swap} has comparatively \emph{small} $\mathrm{RAD}$ and becomes almost benign at 14B, while \textbf{Logic Reversal} transitions from a sizeable drop at 0.5B to nearly no drop (or even slight improvements) at 14B.
\end{itemize}
These results show that, on the simpler \textsc{MBPP+} benchmark, rewrite brittleness is not uniform across edit types: scaling primarily reduces vulnerability to Constraint Change and Data Transformation, while simple component/logic swaps become much easier for larger models to handle.

\begin{table*}[t]
\centering
\small
\begingroup
\setlength{\tabcolsep}{3pt} 

\resizebox{\textwidth}{!}{%
\begin{tabular}{lrrrr@{\hspace{8pt}}rrrr}
\toprule
& \multicolumn{4}{c}{qwen-2.5-coder-0.5b-instruct} 
& \multicolumn{4}{c}{qwen-2.5-coder-14b-instruct} \\
\cmidrule(lr){2-5} \cmidrule(lr){6-9}
Logic Change Type 
& Total & $\text{Acc}(\mathcal{T}_{\text{ori}})$ ($\uparrow$) & $\text{Acc}(\mathcal{T}_{\text{rew}})$ ($\uparrow$) & $\text{RAD}$ ($\downarrow$) 
& Total & $\text{Acc}(\mathcal{T}_{\text{ori}})$ ($\uparrow$) & $\text{Acc}(\mathcal{T}_{\text{rew}})$ ($\uparrow$) & $\text{RAD}$ ($\downarrow$) \\
\midrule
Component Swap        & 393 & 13.5\% &  6.4\% & 52.8\%
                       & 393 & 46.8\% & 29.3\% & 37.5\% \\
Constraint Change     & 185 &  8.6\% &  4.9\% & 43.8\%
                       & 185 & 50.8\% & 33.5\% & 34.0\% \\
Data Transformation   & 385 & 10.6\% &  3.4\% & \textcolor{High}{68.3\%}
                       & 385 & 46.8\% & 26.8\% & 42.8\% \\
Logic Reversal        &  50 &  8.0\% &  6.0\% & 25.0\%
                       &  50 & 46.0\% & 34.0\% & 26.1\% \\
Workflow Modification &  50 & 10.0\% &  6.0\% & 40.0\%
                       &  50 & 42.0\% & 24.0\% & \textcolor{High}{42.9\%} \\
\midrule
\textbf{HMI (all nine types)} 
& \textbf{--} & \textbf{--} & \textbf{--} & \textbf{0.1778}
& \textbf{--} & \textbf{--} & \textbf{--} & \textbf{0.1237}\\
& \multicolumn{4}{c}{qwen-2.5-0.5b-instruct} 
& \multicolumn{4}{c}{qwen-2.5-14b-instruct} \\
\cmidrule(lr){2-5} \cmidrule(lr){6-9}
Logic Change Type 
& Total & $\text{Acc}(\mathcal{T}_{\text{ori}})$ ($\uparrow$) & $\text{Acc}(\mathcal{T}_{\text{rew}})$ ($\uparrow$) & $\text{RAD}$ ($\downarrow$) 
& Total & $\text{Acc}(\mathcal{T}_{\text{ori}})$ ($\uparrow$) & $\text{Acc}(\mathcal{T}_{\text{rew}})$ ($\uparrow$) & $\text{RAD}$ ($\downarrow$) \\
\midrule
Component Swap        & 393 & 12.0\% &  4.8\% & 59.6\%
                       & 393 & 38.2\% & 26.7\% & 30.0\% \\
Constraint Change     & 185 & 11.9\% &  2.7\% & \textcolor{High}{77.3\%}
                       & 185 & 40.5\% & 27.0\% & 33.3\% \\
Data Transformation   & 385 &  8.3\% &  2.1\% & 75.0\%
                       & 385 & 39.7\% & 21.8\% & \textcolor{High}{45.1\%} \\
Logic Reversal        &  50 &  8.0\% &  2.0\% & 75.0\%
                       &  50 & 44.0\% & 28.0\% & 36.4\% \\
Workflow Modification &  50 &  8.0\% &  2.0\% & 75.0\%
                       &  50 & 30.0\% & 18.0\% & 40.0\% \\
\midrule
\textbf{HMI (all nine types)} 
& \textbf{--} & \textbf{--} & \textbf{--} & \textbf{0.1740}
& \textbf{--} & \textbf{--} & \textbf{--} & \textbf{0.0841}\\
\bottomrule
\end{tabular}
}
\endgroup

\caption{Comparison of logic change type performance between four Qwen-2.5 variants on \texttt{BigCodeBench}. We omit logic change types with fewer than 50 samples to avoid statistical bias. RAD highlighted in \textcolor{High}{red} denotes the highest drop.}
\label{tab:qwen-bigcodebench-logic-types}
\end{table*}

\paragraph{Per-type RAD on BigCodeBench.}~\autoref{tab:qwen-bigcodebench-logic-types} reports the same statistics for \textsc{BigCodeBench}. Compared to \textsc{MBPP+}, all major logic-change types now induce much larger drops:
\begin{itemize}[leftmargin=*]
    \item At 0.5B, \textbf{Constraint Change}, \textbf{Data Transformation}, \textbf{Logic Reversal}, and \textbf{Workflow Modification} all yield extremely high $\mathrm{RAD}$ (often above 70\% for Qwen-2.5-0.5B-Instruct), confirming that models struggle when rewrites modify global data flow or constraints on this more challenging benchmark.
    \item Scaling to 14B substantially improves $\text{Acc}(\mathcal{T}_{\text{ori}})$, but $\mathrm{RAD}$ \emph{remains high} for Data Transformation, Constraint Change, and Workflow Modification, with per-type drops still in the 30--45\% range.
\end{itemize}
Thus, even at larger scales, \textsc{BigCodeBench} exposes persistent brittleness when the rewrite changes how data are transformed, rather than merely swapping local components.

\section{Examples of Regressed Tasks}
\label{appendix:regressed}

We randomly selected 5 tasks from each of MBPP+ and BigCodeBench that PASSED in original but FAILED in code\_rewriting from the evaluation results in Qwen2.5-Coder-32B-Instruct. For each task, we provide
\begin{itemize}
    \item Original task prompt and its canonical solution
    \item Code\_rewriting task prompt and the rewritten canonical solution
    \item Alignment and Difficulty analysis from GPT-5 to investigate (1) if the rewritten prompt aligns with its rewritten solution; (2) whether the difficulty of rewritten task align with its original version.
\end{itemize}
The following case studies illustrate regressions on rewritten tasks judged to have comparable difficulty. Together with the aggregate human-validation results in Appendix C, they provide evidence against a systematic increase in task difficulty as the primary explanation for the observed degradation.
\begin{figure*}[!ht]
\begin{center}
    \includegraphics[width=1.0\textwidth]{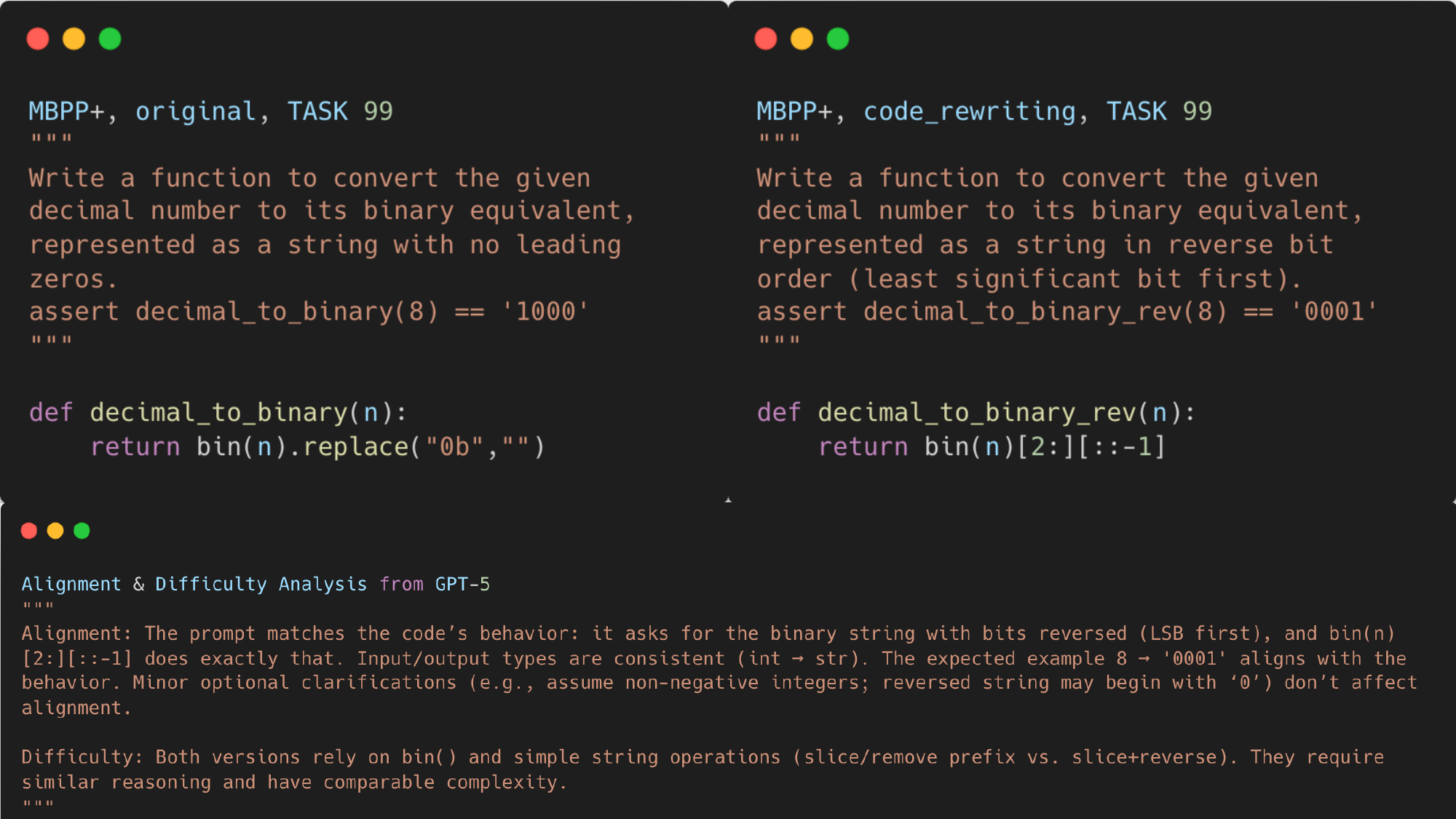}
    \caption{Example of Task-99 from MBPP+ generated from Qwen2.5-Coder-32B-Instruct that PASSED in original but FAILED in code\_rewriting.}
    \label{fig:mbpp_regressed_task_99}
    \vspace{-10pt}
\end{center}
\end{figure*}

\begin{figure*}[!ht]
\begin{center}
    \includegraphics[width=1.0\textwidth]{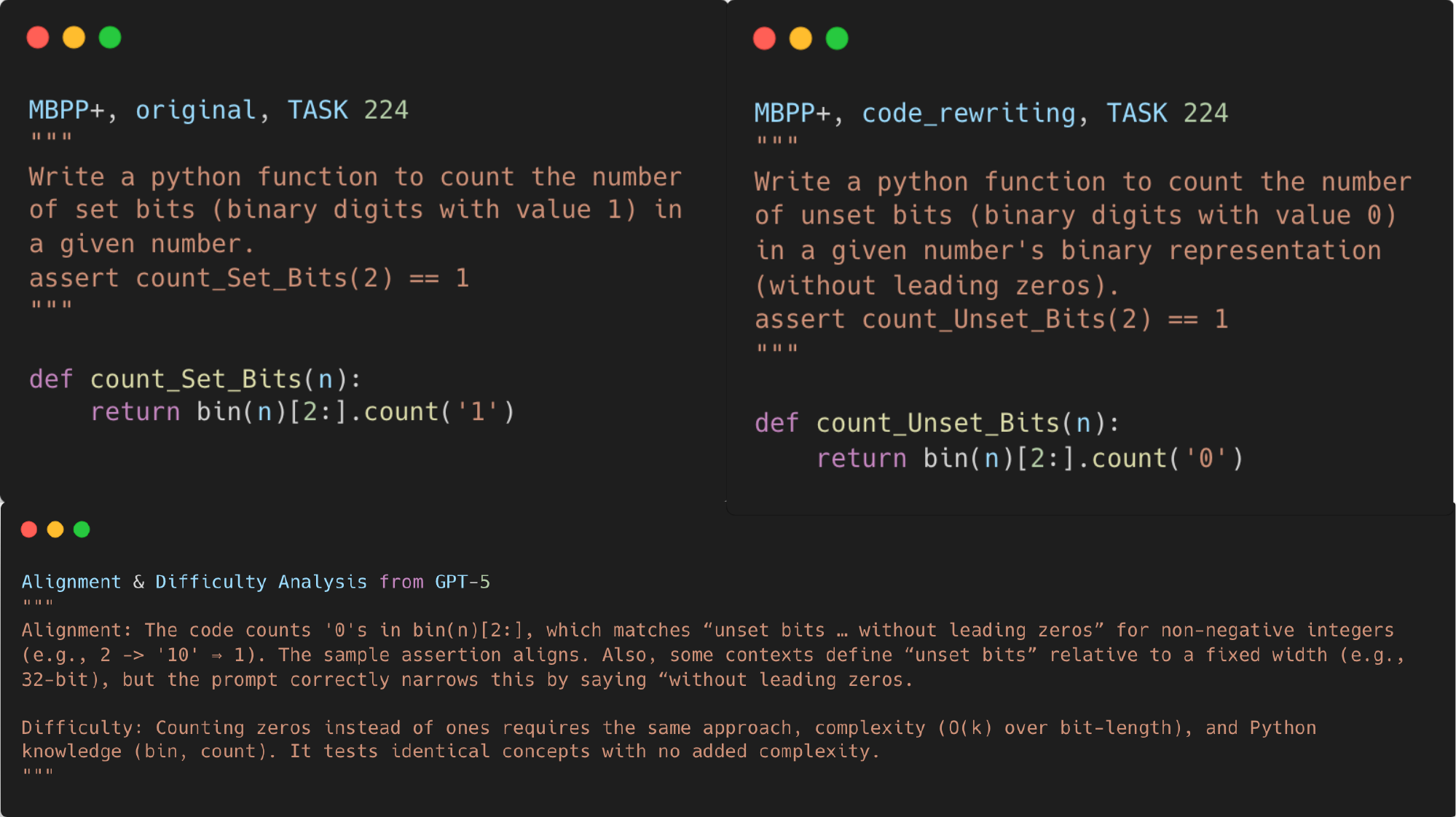}
    \caption{Example of Task-224 from MBPP+ generated from Qwen2.5-Coder-32B-Instruct that PASSED in original but FAILED in code\_rewriting.}
    \label{fig:mbpp_regressed_task_224}
    \vspace{-10pt}
\end{center}
\end{figure*}

\begin{figure*}[!ht]
\begin{center}
    \includegraphics[width=1.0\textwidth]{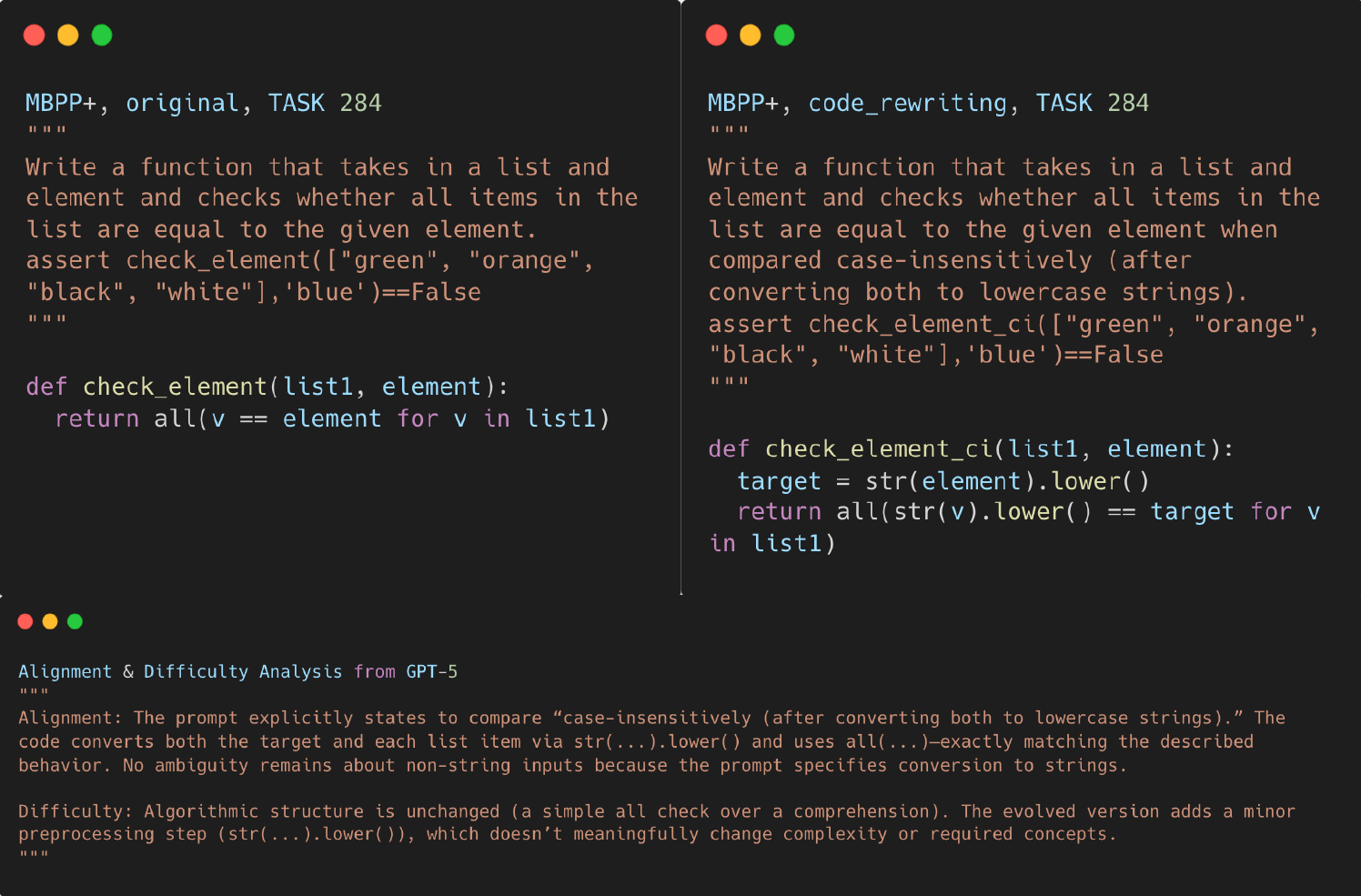}
    \caption{Example of Task-284 from MBPP+ generated from Qwen2.5-Coder-32B-Instruct that PASSED in original but FAILED in code\_rewriting.}
    \label{fig:mbpp_regressed_task_284}
    \vspace{-10pt}
\end{center}
\end{figure*}

\begin{figure*}[!ht]
\begin{center}
    \includegraphics[width=1.0\textwidth]{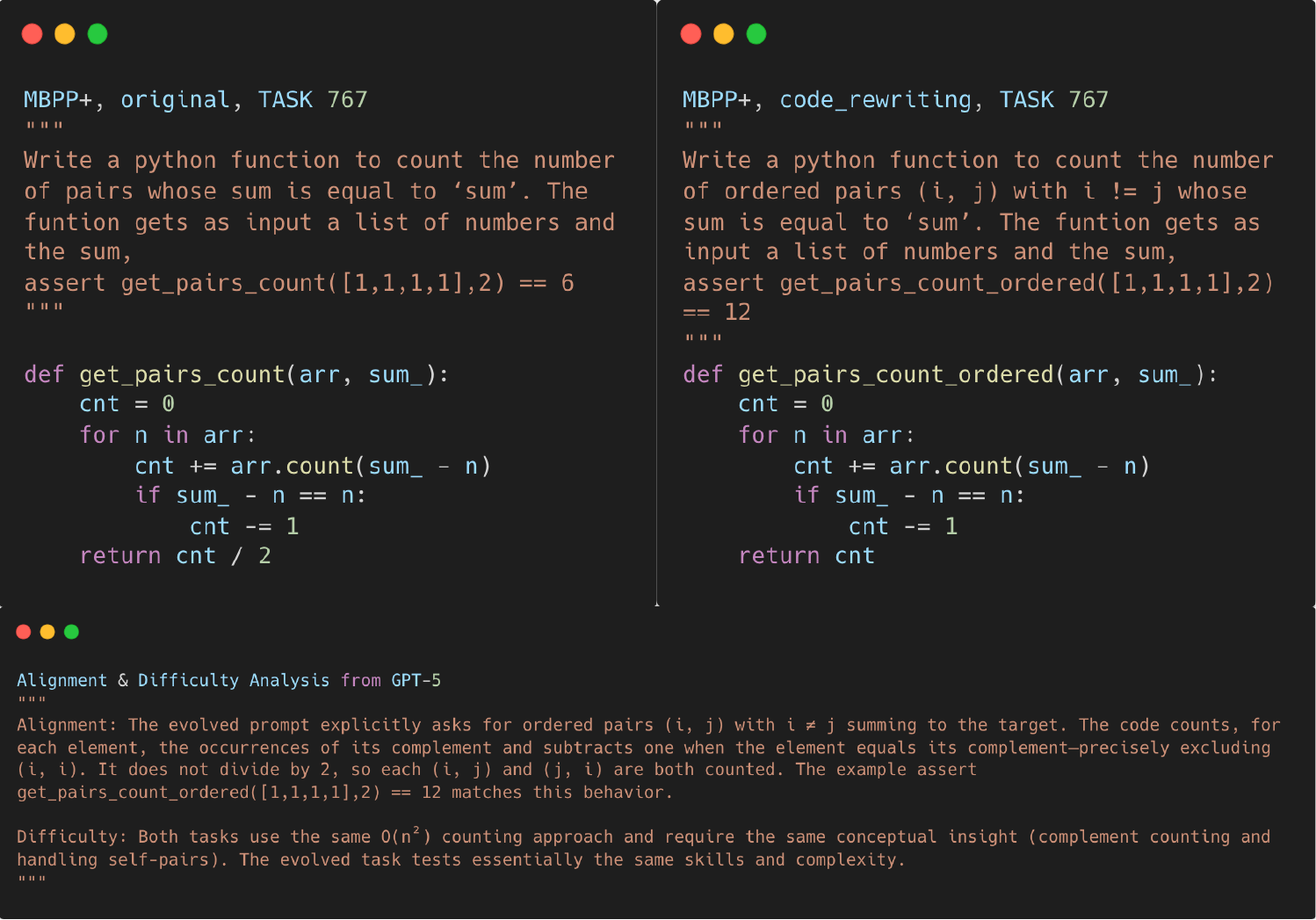}
    \caption{Example of Task-767 from MBPP+ generated from Qwen2.5-Coder-32B-Instruct that PASSED in original but FAILED in code\_rewriting.}
    \label{fig:mbpp_regressed_task_767}
    \vspace{-10pt}
\end{center}
\end{figure*}

\begin{figure*}[!ht]
\begin{center}
    \includegraphics[width=1.0\textwidth]{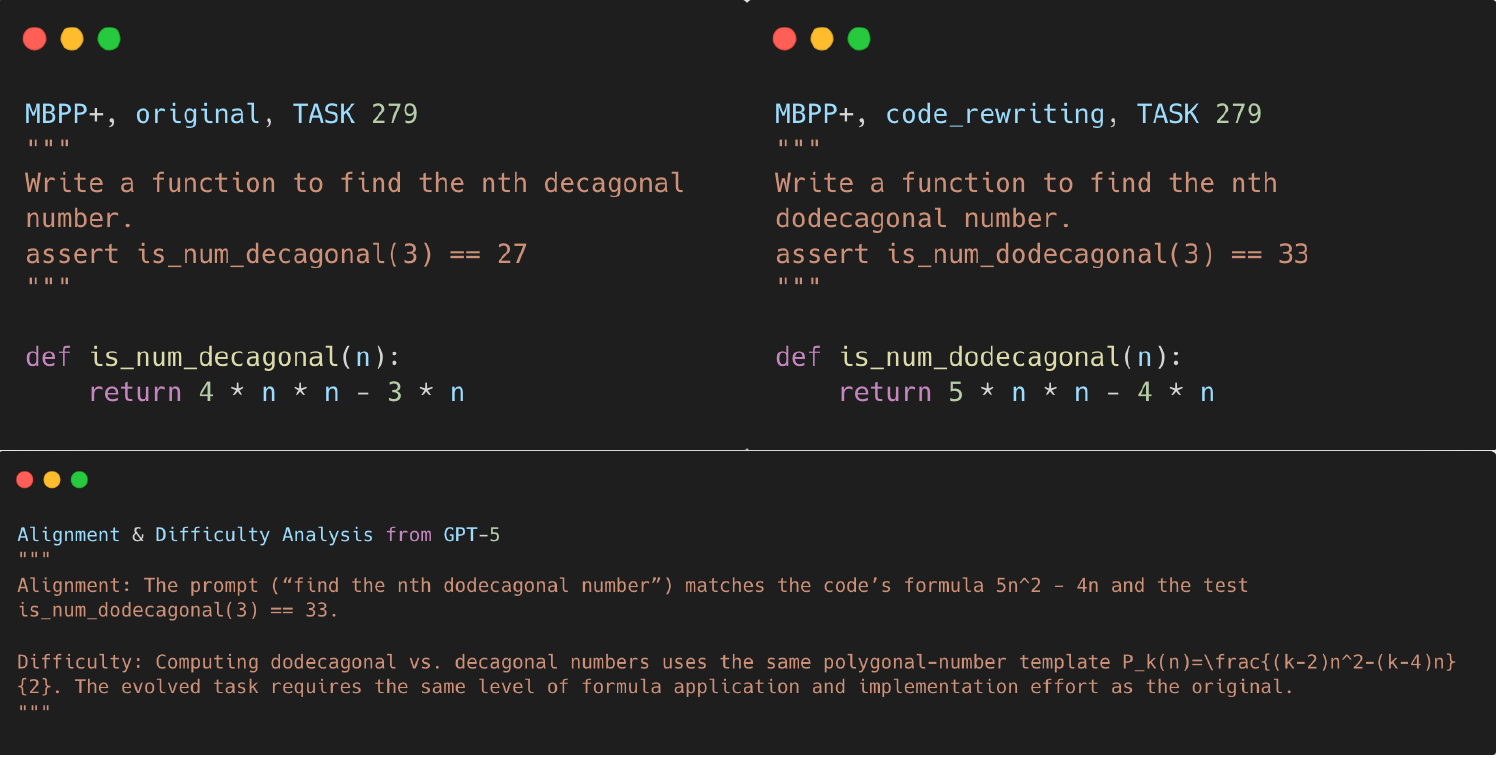}
    \caption{Example of Task-279 from MBPP+ generated from Qwen2.5-Coder-32B-Instruct that PASSED in original but FAILED in code\_rewriting.}
    \label{fig:mbpp_regressed_task_279}
    \vspace{-10pt}
\end{center}
\end{figure*}

\begin{figure*}[!ht]
\begin{center}
    \includegraphics[width=1.0\textwidth]{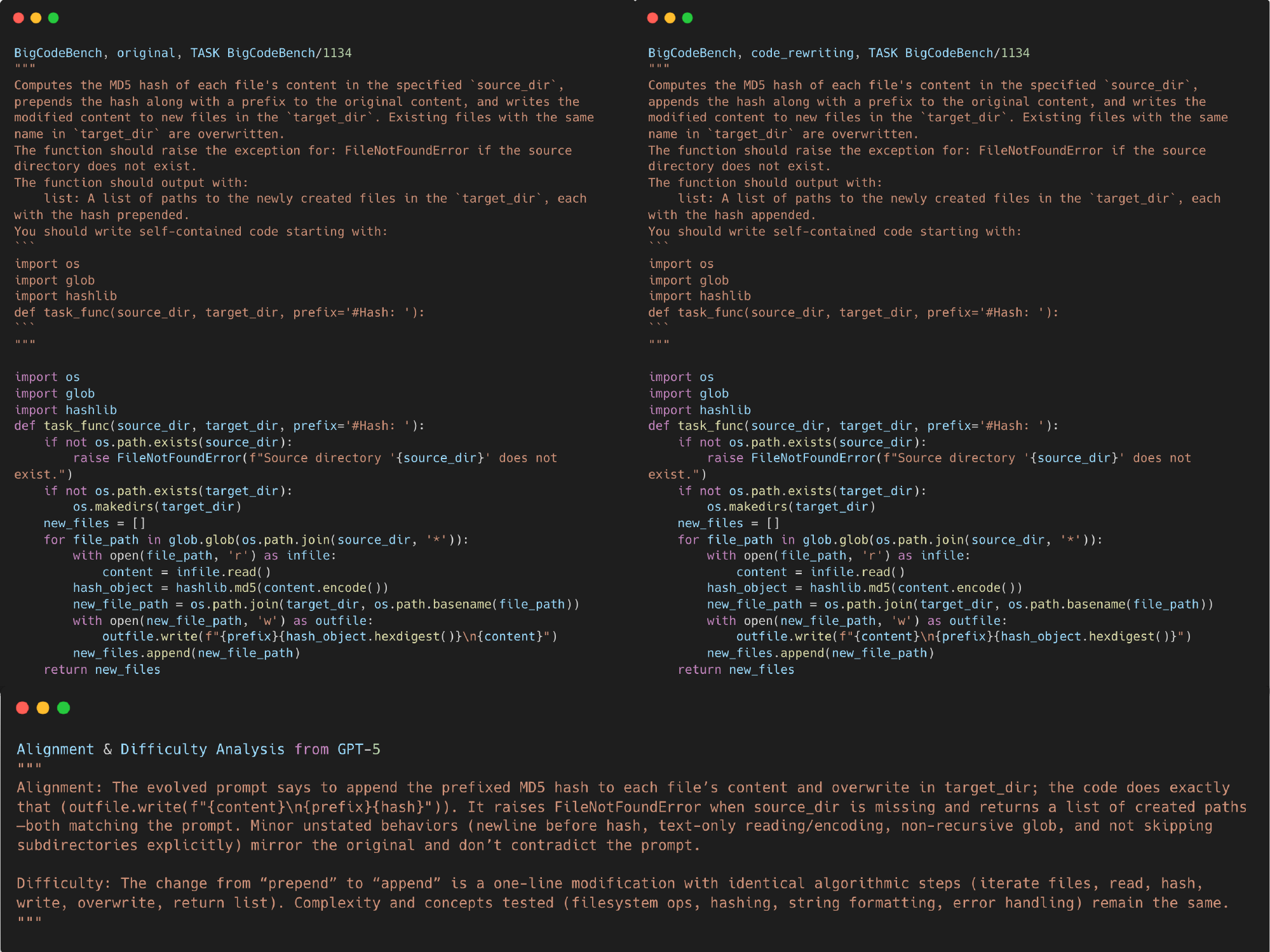}
    \caption{Example of Task-1134 from BigCodeBench generated from Qwen2.5-Coder-32B-Instruct that PASSED in original but FAILED in code\_rewriting.}
    \label{fig:bcb_regressed_task_1134}
    \vspace{-10pt}
\end{center}
\end{figure*}

\begin{figure*}[!ht]
\begin{center}
    \includegraphics[width=1.0\textwidth]{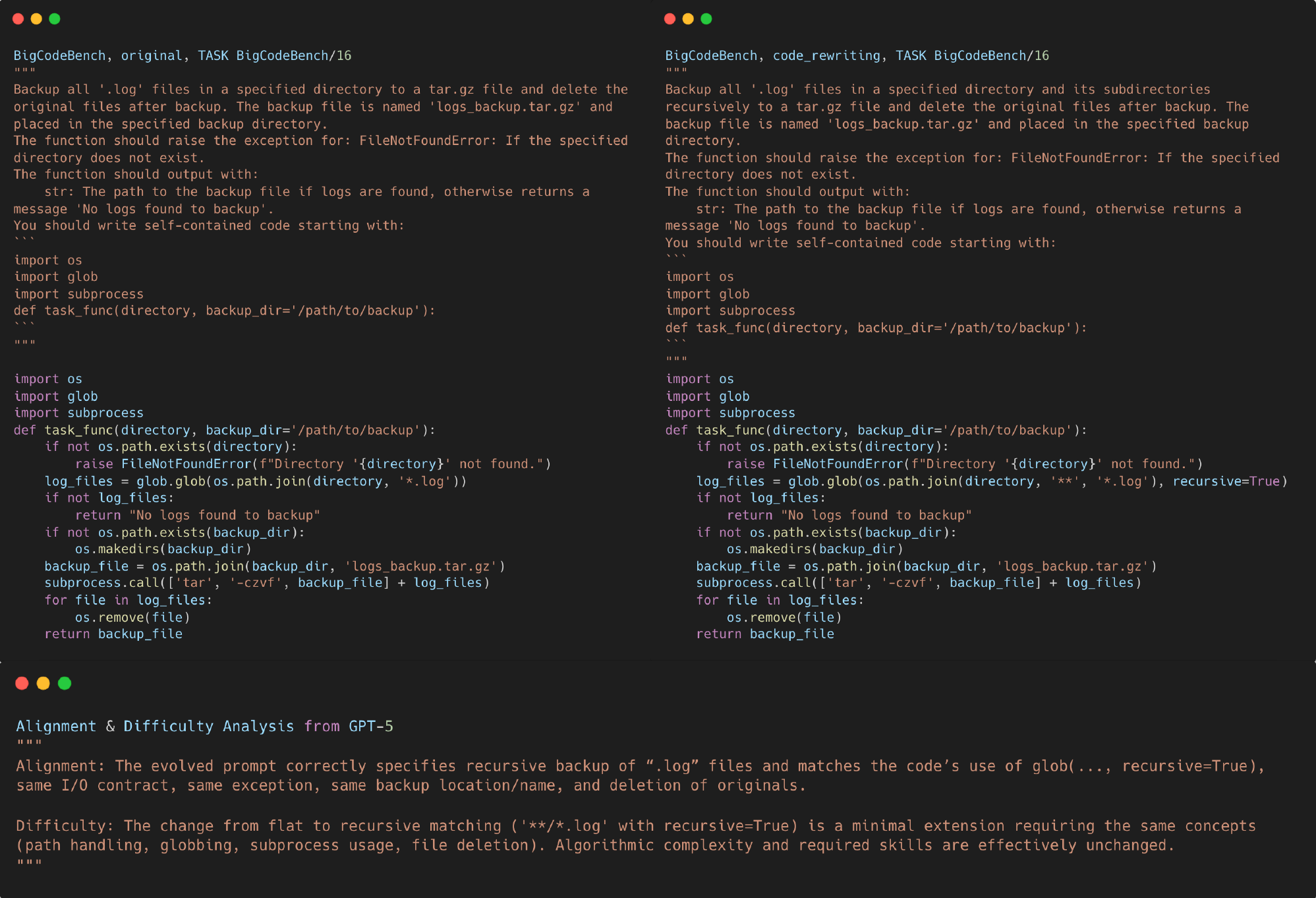}
    \caption{Example of Task-16 from BigCodeBench generated from Qwen2.5-Coder-32B-Instruct that PASSED in original but FAILED in code\_rewriting.}
    \label{fig:bcb_regressed_task_16}
    \vspace{-10pt}
\end{center}
\end{figure*}

\begin{figure*}[!ht]
\begin{center}
    \includegraphics[width=1.0\textwidth]{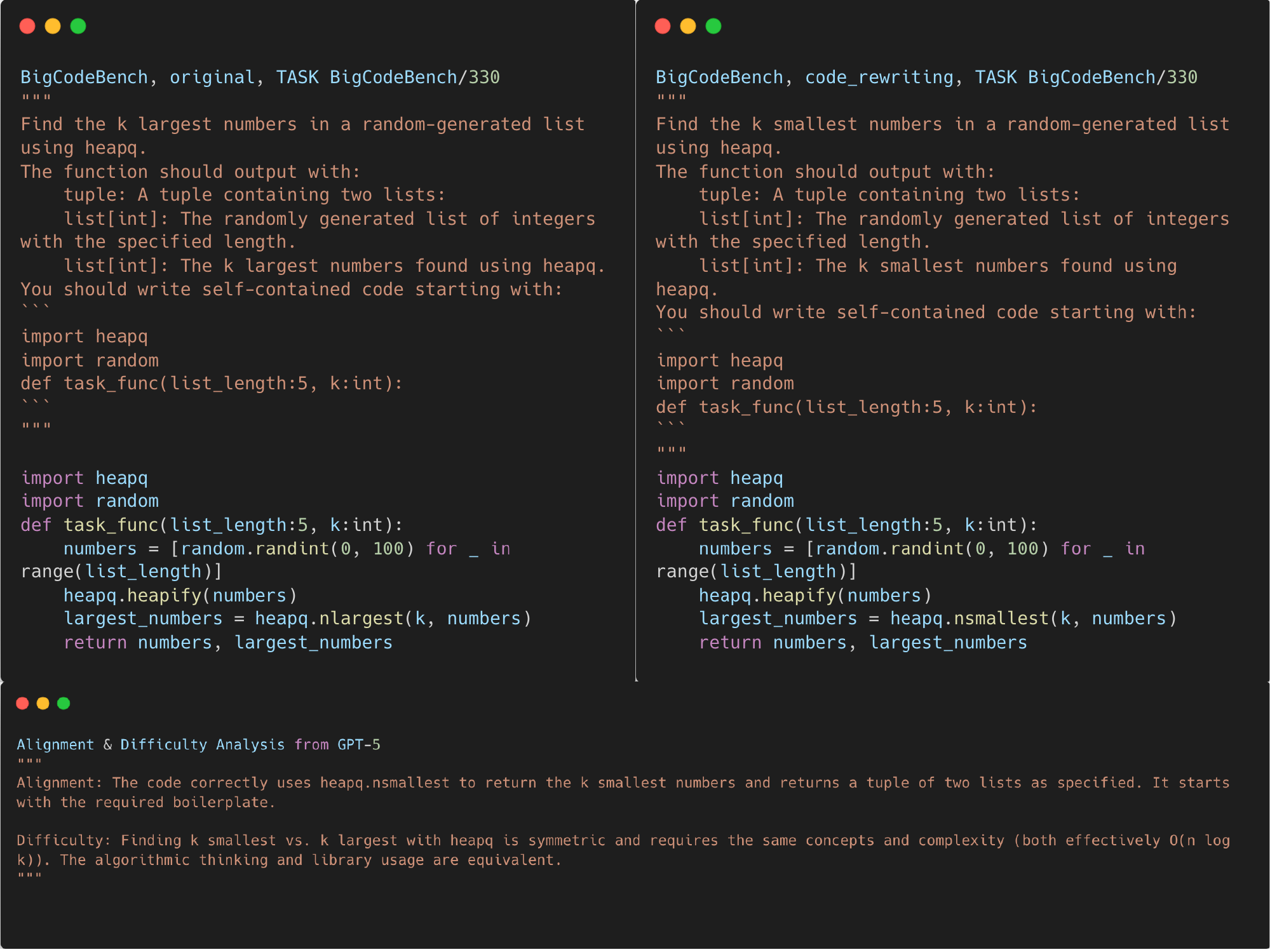}
    \caption{Example of Task-330 from BigCodeBench generated from Qwen2.5-Coder-32B-Instruct that PASSED in original but FAILED in code\_rewriting.}
    \label{fig:bcb_regressed_task_330}
    \vspace{-10pt}
\end{center}
\end{figure*}

\begin{figure*}[!ht]
\begin{center}
    \includegraphics[width=1.0\textwidth]{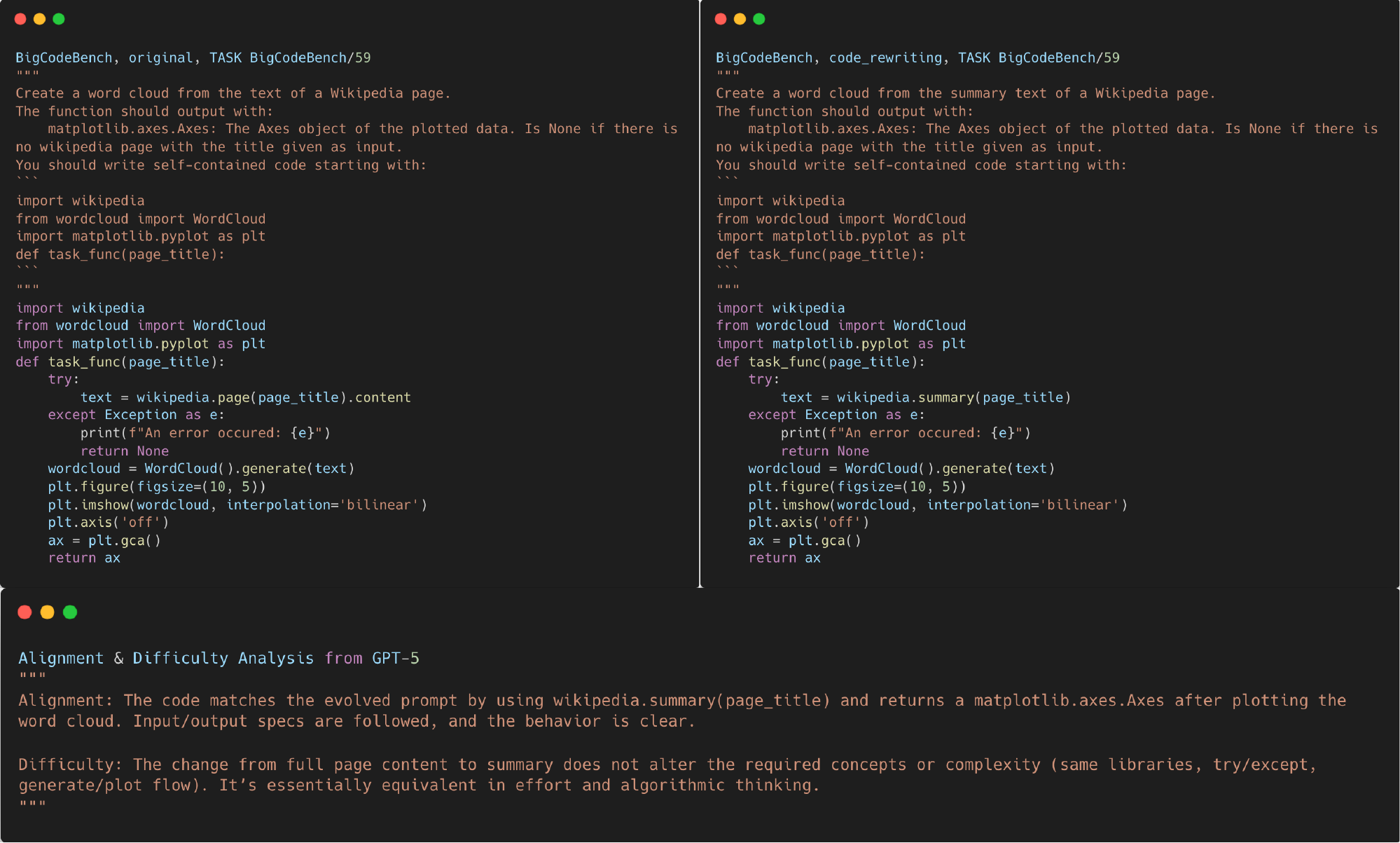}
    \caption{Example of Task-59 from BigCodeBench generated from Qwen2.5-Coder-32B-Instruct that PASSED in original but FAILED in code\_rewriting.}
    \label{fig:bcb_regressed_task_59}
    \vspace{-10pt}
\end{center}
\end{figure*}

\begin{figure*}[!ht]
\begin{center}
    \includegraphics[width=1.0\textwidth]{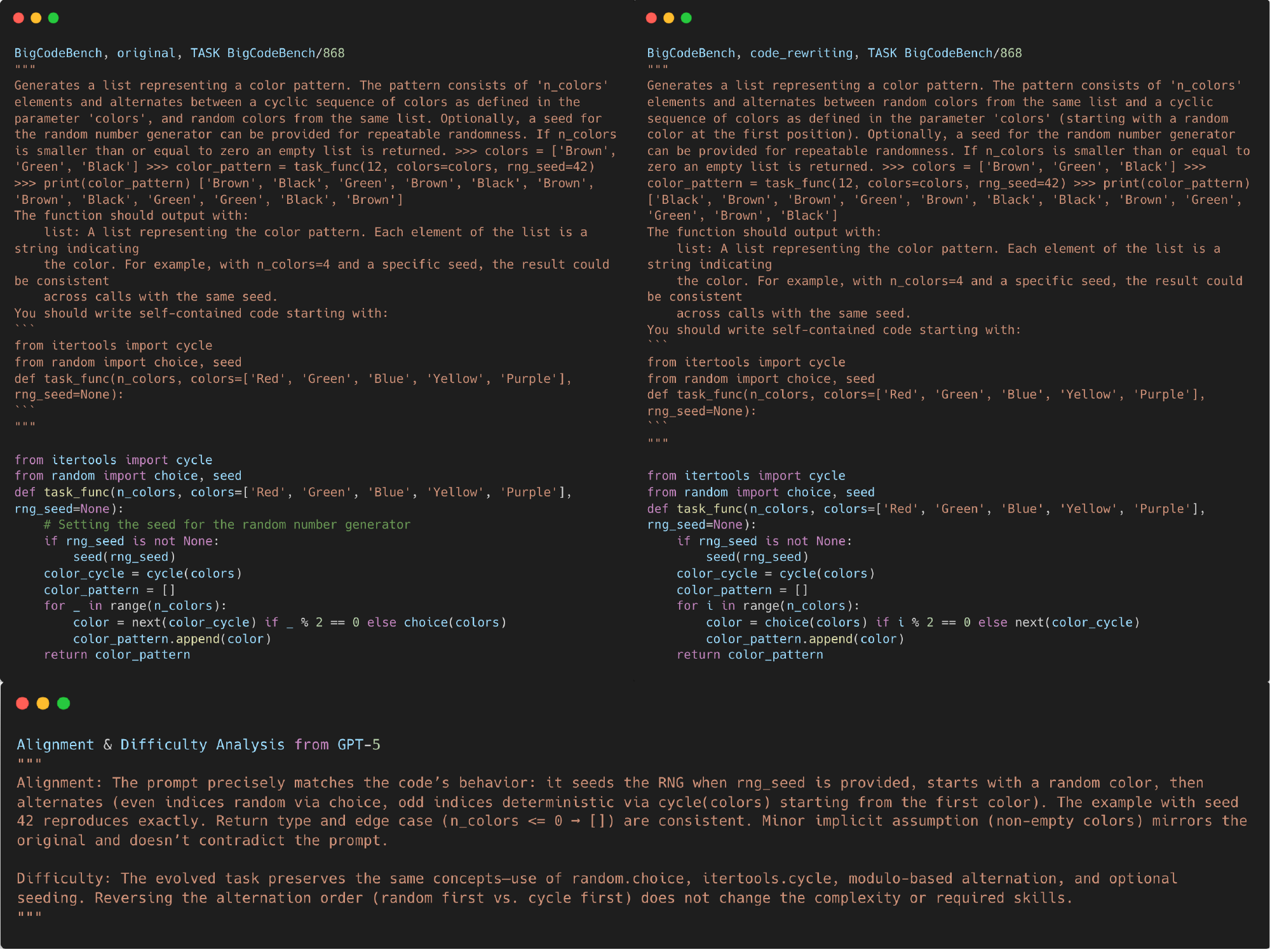}
    \caption{Example of Task-868 from BigCodeBench generated from Qwen2.5-Coder-32B-Instruct that PASSED in original but FAILED in code\_rewriting.}
    \label{fig:bcb_regressed_task_868}
    \vspace{-10pt}
\end{center}
\end{figure*}

\end{document}